\documentclass[pdflatex,sn-basic,iicol]{sn-jnl}

\usepackage{graphicx}%
\usepackage{multirow}%
\usepackage{amsmath,amssymb,amsfonts}%
\usepackage{amsthm}%
\usepackage{mathrsfs}%
\usepackage[title]{appendix}%
\usepackage{xcolor}%
\usepackage{textcomp}%
\usepackage{manyfoot}%
\usepackage{booktabs}%
\usepackage{algorithm}%
\usepackage{algorithmicx}%
\usepackage{algpseudocode}%
\usepackage{listings}%
\usepackage[table]{xcolor}%
\usepackage{cleveref}
\usepackage{algorithm}
\usepackage{algpseudocode}
\algrenewcommand\algorithmicrequire{\textbf{Input:}}
\usepackage{bm}
\usepackage{subcaption}
\usepackage{array}
\usepackage{colortbl}
\usepackage{adjustbox}

\theoremstyle{thmstyleone}%
\theoremstyle{thmstyletwo}%

\theoremstyle{thmstylethree}%

\begin{document}

\title[Article Title]{	
P-PatchDiff: Progressive Patch Diffusion Models for Low-light Image Enhancement}


\author[1]{\fnm{Ruoyu} \sur{Guo}}\email{ruoyu.guo@student.unsw.edu.au}

\author[1]{\fnm{Haonan} \sur{Zhong}}\email{h.zhong.1@unsw.edu.au}

\author[1]{\fnm{Maurice} \sur{Pagnucco}}\email{morri@cse.unsw.edu.au}

\author*[1]{\fnm{Yang} \sur{Song}}\email{yang.song1@unsw.edu.au}

\affil[1]{\orgdiv{School of Computer Science and Engineering}, \orgname{University of New South Wales}, \orgaddress{\city{Sydney}, \country{Australia}}}


\abstract{Recent advancements in low-light image enhancement have leveraged diffusion models for their strong ability to generate perceptually realistic, detailed images. Patch diffusion models further offer a promising solution to size-agnostic image restoration while improving efficiency. However, existing methods typically rely on small, fixed patches (e.g., 64$\times$64) that cannot capture image-level brightness context, whereas enlarging the receptive field improves brightness and colour estimation but substantially increases computational cost. Moreover, low-light images often exhibit uneven brightness across regions,  making it necessary to ensure that locally enhanced patches remain visually coherent when combined into the full image. To address these limitations, we propose \textbf{P-PatchDiff}, a scalable progressive patch diffusion framework for low-light image enhancement that dynamically adjusts patch size throughout the denoising process, enabling a gradual shift from local to global views. A \textbf{Multi-Patch Alignment} strategy is also introduced to normalise features across varying patch scales  using an estimated global brightness proxy. Rather than pursuing pixel-level reconstruction accuracy,  P-PatchDiff focuses on scalability and coherent brightness across the whole image, allowing the model to perceive multi-scale information and better enhance regions with varying brightness. We empirically demonstrate that P-PatchDiff effectively enhances images ranging from 400 $\times$ 600 to 4K and is \textbf{80$\times$} faster than existing patch diffusion models while using less than \textbf{9GB} of memory. The code is available at \url{https://github.com/RuoyuGuo/P-PatchDiff}.}

\keywords{Low-light image enhancement, Patch diffusion model, Denoising diffusion model, Image restoration}



\maketitle

\section{Introduction}\label{sec1}

Low-light image enhancement is challenging due to diverse degradations that obscure fine details, reduce visibility, and limit available information, complicating the enhancement process. Therefore, it is necessary to develop sophisticated enhancement methods with strong generative capabilities. 

Recent advancements in deep learning have enabled substantial improvements in low-light image enhancement by training on real-world paired datasets~\citep{RetinexNet,URetinexNet,Restormer,SNR,Retinexformer}. However, these deep learning methods typically rely on regression models with pixel-level losses (e.g., $\mathcal{L}1$ and $\mathcal{L}2$), resulting in ``averaged'' outputs. While such models often achieve high peak signal-to-noise ratio (PSNR) scores, they struggle to restore missing structures and tend to produce overly smooth images. In contrast, diffusion models, which are likelihood-based, offer a stable training process and generate images with greater detail, making them increasingly popular for low-light enhancement tasks~\citep{pydiff,diffretinex,GSAD}.

\begin{figure}[t]
\centering
\includegraphics[width=1\linewidth]{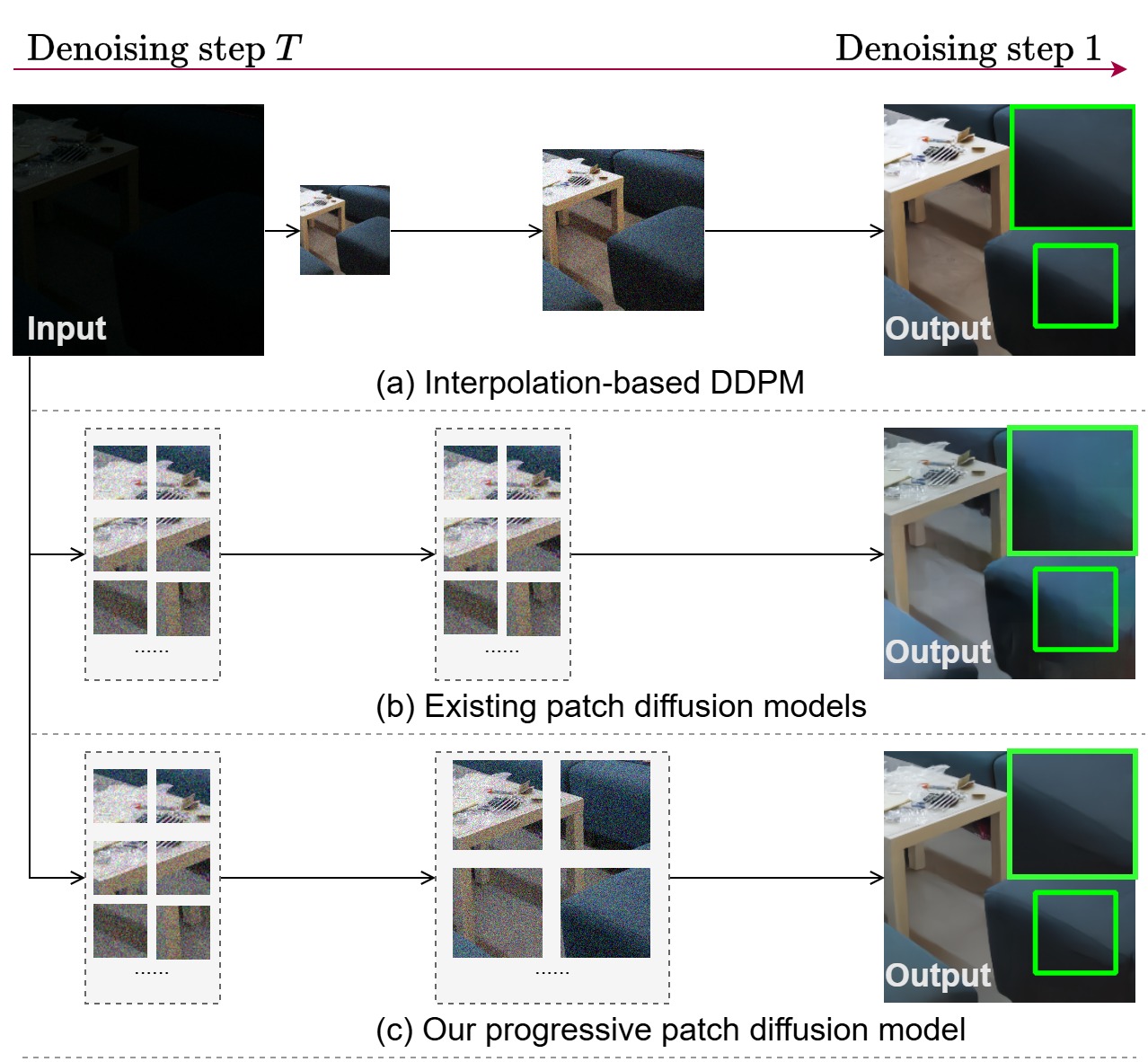}
\caption{Visual comparisons of recent state-of-the-art efficient DDPMs for low-light image enhancement. (a) PyDiff~\citep{pydiff} applies interpolation to resize images, resulting in colour shifting. (b) WeatherDiff~\citep{weatherdiff} uses fixed patch sizes, leading to boundary artefacts. (c) Our method progressively increases patch sizes, producing fine-detail enhanced images.}
\label{fig:teaser}
\end{figure}

Training denoising diffusion probabilistic models (DDPMs) is computationally expensive~\citep{patchdiffusion}. Additionally, modern devices photograph images at varying high resolutions~\citep{lsrw,uhdlol}, increasing the need for efficient frameworks. PyDiff mitigates this by applying interpolation to noise images to reduce sizes during sampling~\citep{pydiff}. A drawback is that it still operates on the full image later, which requires significant memory for high-resolution images. Additionally, as shown in \Cref{fig:teaser} (a), interpolation can cause colour-shifting issues. To address these issues, patch diffusion models have gained attention for operating on patches instead of full images, providing scalability for arbitrary sizes while improving memory efficiency and training speed~\citep{weatherdiff,patchdiffusion}. 

However, they typically use small fixed patch sizes (e.g., $64 \times 64$) for efficiency, but this limits their ability to capture multi-level information and results in boundary artefacts, as illustrated in \Cref{fig:teaser} (b). Additionally, low-light images often exhibit spatially variant illumination, where bright and dark regions coexist within the same image due to exposure imbalance or local light sources~\citep{he2023low,lv2021attention}. This characteristic makes low-light enhancement fundamentally different from uniform illumination adjustment, as it requires a careful balance between global brightness correction and local dark-region enhancement. Prior studies try to utilise multi-scale patches~\citep{MDMS,MIRNet,mei2023pyramid} to address this issue.  However, the lack of a global reference often leads to moderate brightness inconsistency. Moreover, ensembling multi-scale large patches poses two additional challenges. First, it significantly increases the number of patches per denoising step, thus greatly increasing sampling time. Secondly, as shown in \Cref{fig:patchgaps}, our experiments reveal that different patch scales contain varying information, such as colour and brightness, thus simply averaging outputs from multi-scale patches like previous methods would generate suboptimal images~\citep{MDMS}.

Some other methods attempt to incorporate a global corrector~\citep{pydiff,clediff} to achieve balanced illumination enhancement, but relying solely on global information often overwrites local intensity variations. Consequently, models that depend only on multi-scale information or purely global cues struggle to effectively enhance complex, spatially different low-light images. Unfortunately, existing approaches rarely integrate both functionalities in a unified manner. Given that some low-light benchmarks~\citep{DICM,VV} include a range of large-size images with complex lighting environments and that diffusion models typically require long training times, exploring patch diffusion for low-light image enhancement remains valuable. Therefore, in this paper, we aim to tackle two challenging problems in low-light image enhancement simultaneously:  balancing local dark-region enhancement with globally coherent brightness, as well as making patch diffusion models scalable across different GPU devices without degrading image quality.

To this end, we propose a \textbf{P}rogressive \textbf{Patch} \textbf{Diff}usion model (\textbf{P-PatchDiff}) for low-light image enhancement. P-PatchDiff provides a single, integrated solution that tackles both aforementioned challenges at once. We seamlessly incorporate multi-level information by progressively increasing the patch sizes based on the time step. Thus, P-PatchDiff can focus on each local region without interference from other regions. Moreover, we propose a multi-patch alignment method that normalises features across different patch scales using an estimated global brightness proxy, ensuring that features are gradually aligned at each level while capturing global context. To improve the efficiency and scalability of P-PatchDiff, we first investigate the impacts of patch size and stride on enhancement quality and computational overhead. We empirically demonstrate that large strides have minimal impact on sampled image quality but significantly reduce inference time. Based on this insight, we propose to increase the stride as the time step increases to reduce the number of cropped patches and inference complexity. To ensure that patch size and stride increase in a coordinated manner, we establish a unified scheduling scheme that jointly controls their growth across time steps, maximising efficiency without compromising enhancement quality. Our P-PatchDiff is orthogonal to previous low-light enhancement methods that model the physical properties of low-light imaging~\citep{RetinexNet,ZeroDCE,guo2025exploring}: (1) we propose a scalable patch diffusion model to balance enhancement at both global and local levels; (2) this progressive patch diffusion model also mitigates the computational overhead of diffusion models, which was overlooked in previous studies.

We conduct extensive experiments on 9 widely used real-world datasets and 1 synthetic dataset, covering image sizes ranging from $400 \times 600$ to 4K captured by different devices and exposure levels, to validate the effectiveness and efficiency of P-PatchDiff. We use in-domain and out-of-domain validation to demonstrate its capacity across different image scales. Without modifying network architectures, P-PatchDiff achieves competitive performance on 4K images while being \textbf{80$\times$} faster than existing patch diffusion models, requiring only \textbf{8.8GB of GPU memory}. Our contributions can be summarised as follows:

\begin{itemize}
\item[--] We propose \textbf{P-PatchDiff}, a scalable progressive patch diffusion model for low-light image enhancement, which progressively increases patch size and stride with the time step to capture multi-level information while maintaining low inference time.
\item[--] We propose a multi-patch alignment method that normalises features across different patch scales to  ensure consistent global brightness among patches of varying sizes.  
\item[--] We empirically demonstrate that P-PatchDiff not only reduces computational cost but also maintains strong performance across low-light image datasets with varying image sizes, outperforming existing patch diffusion models in both efficiency and robustness to input resolution.
\end{itemize}

\section{Related work}\label{sec2}

\subsection{Low-light image enhancement} Low-light image enhancement has been extensively studied, yet it remains a challenging task. Traditional algorithms either stretch image contrast~\citep{GC,DICM} or enhance illumination~\citep{LIME,VV,NPE,MEF,retinextheory}. Recent deep learning-based methods can be grouped into three main categories. The first focuses on designing network architectures with high capacity to learn the transformation from low-light images to normal-light ones~\citep{Retinexformer,Restormer,MIRNet,BREAD,URetinexNet,ultralol, lv2018mbllen,LLFlow,GLADNet}. Recognising the difficulty of directly learning this mapping, recent research has also introduced two alternative approaches: one provides additional guidance to the networks~\citep{SNR,wang2023low,liu2023low,wu2023learning,smg}, while the other decomposes images into multiple components for individual enhancement~\citep{wang2023brighten,DCCN,zhang2021beyond,KinD,yang2020fidelity,fu2023you,xu2020learning}. 

Moreover, unsupervised methods have been proposed for data-efficient training~\citep{ZeroDCE,ZeroDCE++,PairLIE,RUAS,yang2023implicit,SCI,saini2024specularity,zeroig,guo2025exploring}. Some recent studies have introduced alternative colour spaces for image representation to reduce noise amplification compared to conventional spaces~\citep{Yan_2025_CVPR}. Despite their effectiveness, these regression-based methods often produce blurry outputs, reducing the amount of useful information in the enhanced images. Therefore, we propose to leverage the advantages of diffusion models to address the challenges in low-light enhancement.

\subsection{
Diffusion-based image enhancement}
Diffusion models have recently gained attention in image enhancement for their ability to produce high-quality, perceptually aligned results~\citep{platte,SR3,srdiff}. In low-light image enhancement, methods such as PyDiff improve efficiency by applying image interpolation during training and sampling~\citep{pydiff}, while others adapt pre-trained diffusion models for low-light restoration~\citep{BOOST,quadpior}.  Alternatively, diffusion models have
also been employed to suppress noise in a lookup table-based framework~\citep{lin2025dplut}.  However, due to the severe degradations and complex illumination characteristics in low-light images, directly applying a diffusion model often fails to capture structural and tonal variations effectively.

To address these challenges, some approaches first transform the image into alternative representations and then apply diffusion models to enhance specific components. Such transformations include Retinex decomposition~\citep{diffretinex,lightendiff}, wavelet transform~\citep{wavediff}, and Fourier transform~\citep{lv2024fourier,MDMS}. However, traditional decomposition methods generally overlook the degradations present in low-light inputs, leading to suboptimal enhancement~\citep{DCCN}. Furthermore, these approaches typically require separate diffusion models to enhance each component, thereby increasing both complexity and training cost.

Another strategy for improving diffusion-based enhancement is to incorporate prior knowledge into the denoising process. For example, CLE-Diffusion integrates brightness levels as conditioning signals~\citep{clediff}, while other methods embed Retinex priors~\citep{joresdiff,kang2024image}, exposure priors~\citep{Wang_2023_ICCV}, or degradation priors~\citep{lldiff}. More recently, large language models have been used to generate image quality maps that serve as guidance for diffusion models~\citep{zhou2025low}. Motivated by the effectiveness of such priors,  we propose estimating a global brightness proxy from full images and using it to align outputs across different patch scales with the global statistics of the target image.

\subsection{
Patch diffusion models}
Patch diffusion models crop small patches during training to improve computational efficiency~\citep{patchdiffusion,hu2024learning}, with theoretical proof that the noise distribution can be learned from patches rather than full images. These early works also noted the problem of boundary artefacts, although without extensive analysis. Moreover, most still perform sampling on full-size images, which is impractical for image enhancement tasks due to variable image resolutions.

To address this limitation, feature-level patching methods have been proposed. For example, Patch-DM~\citep{ding2023patched} encodes neighbouring overlapping patches and merges them in the feature space, enabling high-resolution sampling from patches. However, this requires carefully designed position embeddings to maintain spatial consistency, which increases model complexity. An alternative is hierarchical patch diffusion~\citep{skorokhodov2024hierarchical,hur2025high}, which first generates a small image and then repeatedly performs denoising on progressively larger noise inputs, guided by the previously generated results. While this avoids the need for full-size inputs, it significantly increases time and network complexity due to multiple upsampling stages.

To simplify the inference process, WeatherDiff~\citep{weatherdiff} proposes a mean-estimated noise merging strategy, where overlapping patches are denoised independently and then averaged, mitigating the complexity of feature-level merging or iterative upsampling. However, it uses a fixed small patch size and thus lacks multi-scale context.

MDMS~\citep{MDMS} addresses this by denoising multiple copies of the image cropped at different patch sizes at each denoising step, then averaging the outputs. This multi-scale patching not only provides richer global–local information for enhancement but also reduces boundary artefacts. Similarly, DiffInfinite~\citep{aversa2023diffinfinite} segments each patch into subregions based on semantic masks, then repeatedly denoises each subregion until all are complete, before merging them. While effective for handling complex spatial structures, these approaches result in significantly longer sampling times and do not address colour or brightness inconsistencies between scales. Unlike previous patch diffusion models, our approach crops only one scale of patches per denoising step. Multi-level information is introduced by progressively increasing patch size and stride in a time-aware manner, reducing the large number of patches generated by multi-scale cropping and maintaining high performance. Additionally, a multi-patch alignment method is proposed to mitigate inconsistencies across scales effectively.

\subsection{Progressive strategy}
The progressive strategy is commonly used in image generation to explore multi-scale information~\citep{ren2019progressive,Restormer}. Although some previous works~\citep{skorokhodov2024hierarchical,jiang2020multi,zamir2021multi} employ progressive strategies, they rely on specially designed network architectures, which limit their applicability to broader scenarios.   Additionally, other lines of work~\citep{li2021low,knaus2014progressive} focus on iterative image refinement to improve perceptual quality, without addressing patch-level efficiency or multi-scale interaction. To integrate global information into progressive models, IA-YOLO~\citep{liu2022image} generates low-resolution inputs for CNNs to predict image filter parameters. However, these predicted parameters are used for traditional signal processing, which is less effective.  Overall, while our method builds upon the general idea of progressive processing, it is proposed to address issues of patch diffusion models, offering a novel perspective for low-light image enhancement.

\section{Methods}
\begin{figure*}[t]
\centering
\includegraphics[width=1\linewidth]{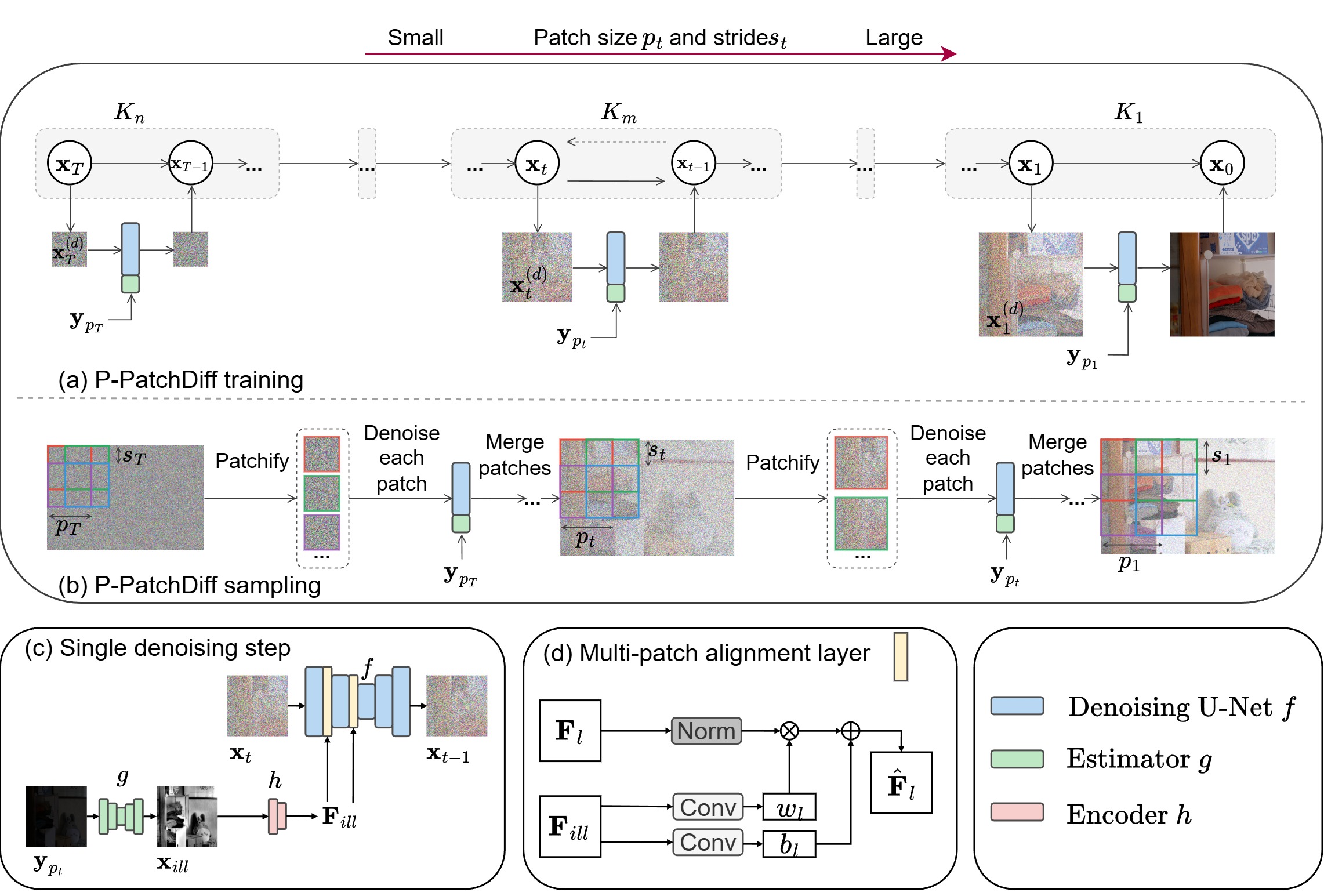}
\caption{Overall framework of P-PatchDiff. We divide the total time steps $T$ into $n$ equal-length subsets $K$.  (a) During training, we randomly crop a patch $\textbf{x}_t^{(i)}$ with size $p_t$, where $1  \le t \le T$, from the normal-light image $\textbf{x}_0$ for diffusion and denoising. The patch size $p_t$ is progressively increased to capture multi-level image content. (b) During sampling, the stride $s_t$ is also progressively increased to reduce the number of patches. (c) We downsample the low-light image to $\textbf{y}_{p_t}$ with a size of $p_t \times p_t$. An estimator $g$ is then utilised to produce  the global brightness proxy $\textbf{x}_{ill}$, which is subsequently processed by the encoder $h$ to extract $\textbf{F}_{ill}$. (d) The structure of the multi-patch alignment layer in $f$. We utilise $\textbf{F}_{ill}$ to normalise the encoder feature $\textbf{F}_l$ and generate the enhanced features $\hat{\textbf{F}}_{l}$.}
\label{fig:overall}
\end{figure*}

As illustrated in \Cref{fig:overall}, our goal is to train a denoising U-Net $f_\theta$ to generate an enhanced image $\tilde{\textbf{x}}$ that is as close as possible to the normal-light image  $\textbf{x}_0$ from a randomly sampled noise image $\textbf{x}_t$, conditioned on the low-light image $\textbf{y}$. We divide the total time steps $T$ into $n$ equal-length subsets $\{K_m\}^{n}_{m=1}$.
Within each subset $K_m$, the patch size $p_t$ and stride $s_t$ are kept fixed, but they change progressively across subsets, guiding the model to transition from local to global enhancement. To improve efficiency, instead of denoising on the full-size noise image $\textbf{x}_t$, we concatenate $\textbf{x}_t$ and $\textbf{y}$, then crop the result into overlapping patches $\textbf{y}^p_s$ using patch size $p_t$ and stride $s_t$. Each $\textbf{y}^p_s$ is denoised independently and then merged to reconstruct the full-size denoised output $\textbf{x}_t$ at time step $t$, where the overlap areas are averaged. 

To address the varying illumination captured by different patch sizes and mitigate the limitation of using image patches instead of the full-size image, we incorporate a lightweight U-Net $g$ to estimate  a global brightness proxy  $\textbf{x}_{ill}$ from a downsampled full-size image $\textbf{y}_{p_t}$, which has a size of $p_t \times p_t$ to align with each patch. Moreover, we encode $\textbf{x}_{ill}$ into a global brightness feature  $\textbf{F}_{ill}$ through an encoder $h$ and incorporate it into $f_{\theta}$ to normalise features across multi-scale patches.

In the following sections, we first introduce patch diffusion models in \Cref{sec:revisiting}, then our progressive patch diffusion model in \Cref{sec:progressive}, and the multi-patch alignment strategy in \Cref{sec:multipatch}.

\subsection{Revisiting patch diffusion models}
\label{sec:revisiting}
In image enhancement tasks, the patch diffusion model (PDM)~\citep{weatherdiff,MDMS,patchdiffusion} trains diffusion models using small local patches of size $p$ cropped from full-size images $\textbf{x}_0$. During sampling, PDM divides the full-size low-light image $\textbf{y}$ into overlapping patches $\textbf{y}^p_s$ using patch size $p$ and stride $s$. Then, $\textbf{y}^p_s$ are concatenated with cropped sampled Gaussian noise to construct the input of $f_\theta$. Finally, these overlapping patches are denoised and then merged to generate the full-size enhanced image $\tilde{\textbf{x}}$, where the overlap areas are averaged~\citep{weatherdiff}.  

Building on this general framework, several patch diffusion models have been developed for low-light image enhancement, each adopting different strategies for patchifying to balance efficiency and quality. Here, we review two representative PDMs: WeatherDiff~\citep{weatherdiff} and MDMS~\citep{MDMS}, 
and analyse how the number of subsets $n$ affects their patch size $p$ and efficiency.

\begin{figure}[t]
\centering
\includegraphics[width=1\linewidth]{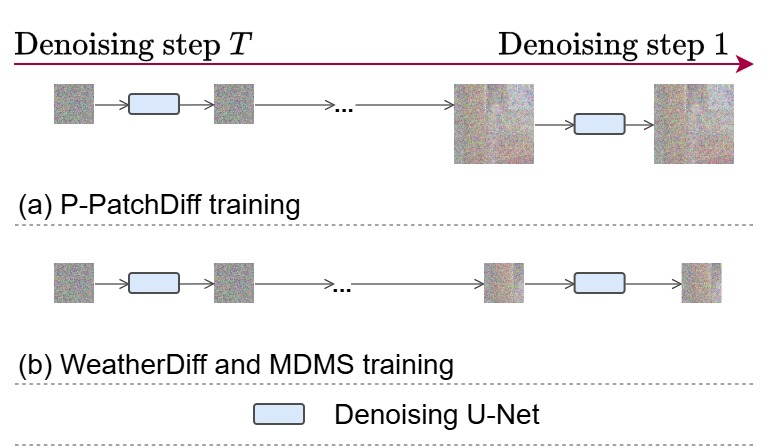}
\caption{Comparison of training strategy between WeatherDiff~\citep{weatherdiff}, MDMS~\citep{MDMS} and our P-PatchDiff. (a) We gradually increase patch sizes during training to enable our denoising models to perceive more information from larger patches. (b) WeatherDiff and MDMS use a fixed patch size throughout all steps during training.}
\label{fig:frametrain}
\end{figure}

\begin{figure}[t]
\centering
\includegraphics[width=1\linewidth]{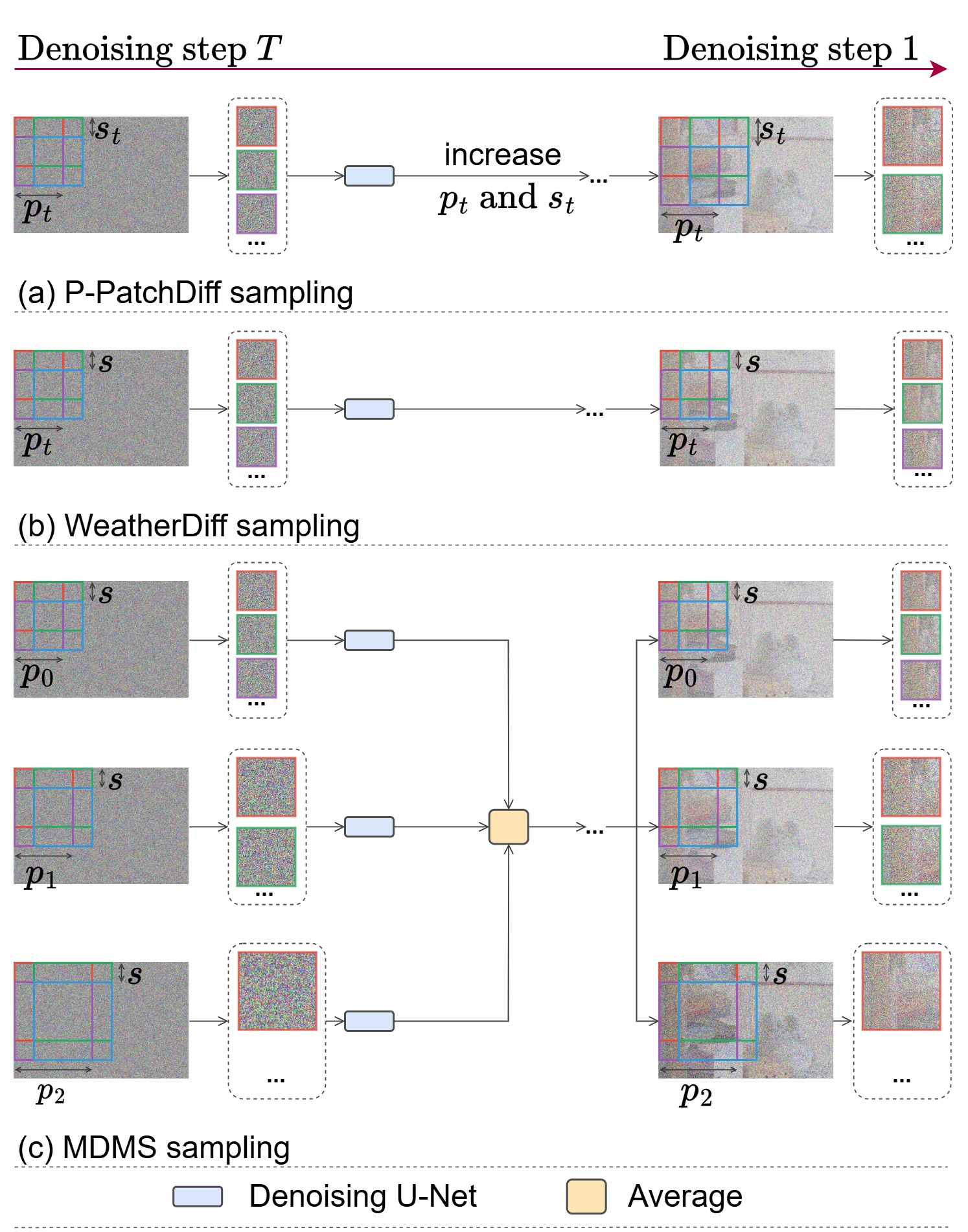}
\caption{Comparison of sampling strategy between WeatherDiff~\citep{weatherdiff}, MDMS~\citep{MDMS} and our P-PatchDiff. (a) We incorporate multi-level information by increasing patch sizes. We also use larger strides for larger patch sizes to reduce the number of patches since the large patches are robust to changes in strides. (b) WeatherDiff 
lacks multi-level information since it only uses a fixed patch size. 
(c) MDMS incorporates multi-level information by creating multiple copies of the same noisy image and cropping each copy with a different patch size, thereby incurring substantial computational costs.}
\label{fig:framesample}
\end{figure}

\textbf{WeatherDiff. } As shown in \Cref{fig:frametrain} (b) and \Cref{fig:framesample} (b), WeatherDiff uses a fixed patch size strategy. Although the timesteps are divided into $n$ subsets, the same patch size 
$p$ is used for all subsets. The value of $p$ is determined by $n$, with a larger $n$ producing a larger patch size:
\begin{equation}
p = 64 + (n - 1) \times 32.
\label{eq:weatherdiff}
\end{equation}

\begin{figure*}[t]
    \centering
    \includegraphics[width=1\linewidth]{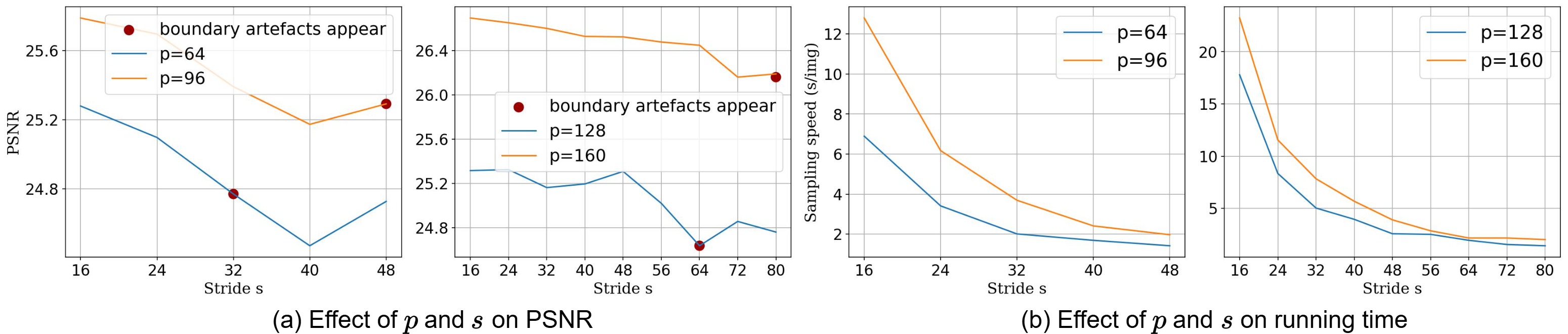}
\caption{Analysis of $p$ and $s$ on enhancing efficiency and effectiveness. Results are obtained by running on the LOL-v2-Real testing set. The red dot indicates the minimum $s$ at which patch boundary artefacts become noticeable. (a) We notice that the patch size can affect enhancement quality. Larger patch sizes are robust to different strides without severe performance drops. (b) Using larger strides can significantly reduce running time.}
\label{fig:psstudy}
\end{figure*}

\begin{figure*}[t]
    \centering
    \includegraphics[width=0.85\linewidth]{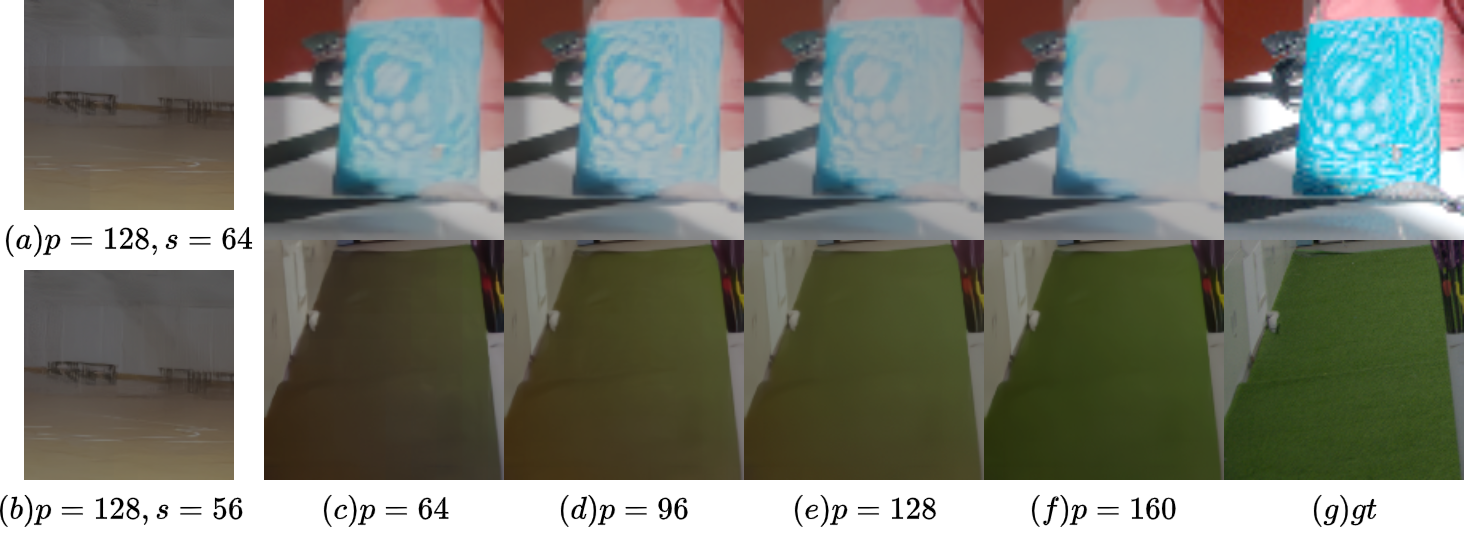}
\caption{(a)-(b): Patch boundary artefacts and inconsistent colour appear around $s=\frac{p}{2}$ but become less visible when $s < \frac{p}{2}$. (c)-(g): Inconsistent enhanced outputs from PDMs trained and evaluated with different $p$ values.}
\label{fig:patchgaps}
\end{figure*}

\noindent This patch size is used for both training and sampling. The full-size image is cropped into overlapping patches of size $p$, which are individually denoised and then merged to form the final output. A larger $n$ leads to a larger $p$, enabling the model to capture more global context but increasing memory and computation cost.

\textbf{MDMS. } As shown in \Cref{fig:frametrain} (b) and \Cref{fig:framesample} (c), MDMS uses a multi-scale patching strategy.
At each denoising step, it creates $n$ copies of $\textbf{x}_t$ and crops each $\textbf{x}_t$ into overlapping patches using different patch sizes, instead of dividing the total time steps into $n$. This generates $n$ sets of patches. The size of the 
$i$-th patch set is defined as:
\begin{equation}
\begin{aligned}
    &p = \{p_1, \dots, p_n\}, \\ 
   &\quad \text{where} \quad p_{1\le i \le n} = 64 + (i - 1) \times 32.
\end{aligned}
\label{eq:mdms}
\end{equation}

\begin{table}[t]
\scriptsize
    \caption{Training time with different $p$ for $500,000$ iterations.}
    \centering
    \begin{tabular}{c|c|c|c|c|c}
    \toprule
        $p$ & 64 & 96 & 128 & 160 & 192 \\
        \midrule
        Time (h) & $\sim 24$  & $\sim 46$ & $\sim 64$ & $\sim 80 $ & $ \sim 98$ \\
        \bottomrule        
    \end{tabular}
    \label{tab:ptraintime}
\end{table}

\noindent Each set is denoised independently with the same method as WeatherDiff, and the output of each set is further averaged to form the denoised output at the current time step. 
When $n=1$, MDMS reduces to WeatherDiff. Equivalently, WeatherDiff can be viewed as a special case of MDMS that uses only the largest patch size $p_n$ instead of multiple scales. Increasing $n$ enhances performance by incorporating multi-scale information, but the number of cropped patch sets and inference passes grows linearly with $n$, leading to much higher computational cost.

These two strategies illustrate the design space of PDMs, ranging from single-scale to multi-scale patch processing.
However, regardless of the specific design, two critical issues limit PDM performance: 1) The choice of $p$ and $s$ significantly impacts the enhanced image quality, since low-light images often exhibit significant variance across patches~\citep{wangill2024}; 2) While MDMS uses multi-scale large patches, which is beneficial for enhancement, it considerably increases sampling times, especially with small $s$. To systematically analyse these effects, we conduct experiments on the LOL-v2-Real dataset~\citep{LOLv2} because it contains a wide range of scenes and multi-exposure images. Specifically, 
following WeatherDiff~\citep{weatherdiff}, we train 4 individual PDMs with $p = 64, 96, 128, \text{and } 160$ on the LOL-v2-Real training set. We then evaluate these PDMs on its testing set with $s$ ranging from $16$ to $80$.

As shown in \Cref{fig:psstudy}, we present the PSNR results across different combinations of $p$ and $s$. We note that  $p$ has a non-trivial impact on enhancement performance, and different $p$-$s$ combinations lead to different trade-offs. In general, increasing $p$ tends to improve performance in terms of PSNR. However, the running time nearly doubles with only a 64-pixel increase in patch size. On the other hand, simply increasing $p$ does not guarantee improved PSNR, as shown by the comparison between $p=96$ and $p=128$ in \Cref{fig:psstudy} (a). This suggests that naively using larger patches is not an optimal strategy for low-light image enhancement, which is further supported by the results in \Cref{tab:abpatch}. Additionally, as shown in \Cref{fig:psstudy} (b), increasing the stride $s$ significantly reduces the running time, but the benefit diminishes once $s$ reaches half of $p$. At the same time, boundary artefacts begin to appear at this point. This indicates that $s = \frac{p}{2}$ is a critical trade-off value.

We also study the impact of $p$ and $s$ on the patch boundary artefacts, which is another issue in PDM-generated images~\citep{MDMS, weatherdiff}. As illustrated in \Cref{fig:patchgaps}, these boundary artefacts become noticeable around $s=\frac{p}{2}$. 
Therefore, to mitigate such artefacts, the stride should satisfy $s < \frac{p}{2}$. Using a larger patch size $p$ consequently allows a larger stride while still meeting this constraint, making larger patches more favourable. Additionally, larger patches better tolerate variations in stride without causing severe performance degradation. However, as shown in \Cref{tab:ptraintime} and \Cref{fig:psstudy}, a significant drawback of using large patches is the substantial increase in training and sampling time. 

Furthermore, as shown in \Cref{fig:patchgaps}, we find that enhanced outputs from PDMs with different $p$ produce variations in colour and illumination. This result indicates that using fixed patch sizes~\citep{weatherdiff} is insufficient to enhance low-light images, and simply averaging patches across scales~\citep{MDMS} cannot fully mitigate the differences between patches. Given these findings, it is essential to develop a strategy to address these challenges in PDMs.

\begin{algorithm}[t]
\scriptsize
\caption{P-PatchDiff Training}\label{alg:ptrain}
\begin{algorithmic}[1]
\Require Normal-light and low-light image pairs $(\mathbf{x}_0, \mathbf{y})$, estimator $g$, brightness proxy encoder $h$, hyperparameters $\Delta s$ and $\Delta p$.
\While{not converged}
\State Sample $t \sim \text{Uniform}\{1, \dots, T\}$ and $\bm{\epsilon}_t \sim \mathcal{N}(\mathbf{0}, \mathbf{I})$
\State Obtain $p_t$ by \Cref{eq:pfun}
\State Generate a binary mask $\mathbf{P}_i$ with size $p_t$
\State Crop $\mathbf{x}_0^{(i)} = \mathbf{P}_i \odot \mathbf{x}_0$, $\mathbf{y}^{(i)} = \mathbf{P}_i \odot \mathbf{y}$, $\mathbf{F}_{ill} = h(g(\mathbf{y}_{p_t}))$
\State Perform a single gradient descent step for
    $\nabla_{\theta}\|\bm{\epsilon}_t - f_\theta(\sqrt{\bar{\alpha}_t}\,\mathbf{x}_0^{(i)} + \sqrt{1-\bar{\alpha}_t}\,\bm{\epsilon}_t, \mathbf{y}^{(i)}, \mathbf{F}_{ill}, t)\|^2$
\EndWhile
\State \textbf{return} $\theta$
\end{algorithmic}
\end{algorithm}

\subsection{Progressive patch diffusion models}
\label{sec:progressive}
Based on the above experiments, we propose P-PatchDiff, a progressive patch diffusion model for low-light image enhancement that dynamically adjusts patch size and strides over time. This adaptive strategy addresses the limitations of fixed PDMs, enabling the model to capture both local and broader contexts effectively while managing computational costs. More importantly, this progressive diffusion could extensively explore local illumination variations within low-light images at each step, rather than treating all regions uniformly. Therefore, it has the potential to more effectively enhance low-light images with partial light sources. As shown in \Cref{fig:overall}, we divide the total of $T$ time steps into $n$ uniform subsets of length $\Delta K$:
\begin{equation}
\begin{aligned}
\{1, 2, \dots, T\} &= \bigcup_{m=1}^{n} K_m, \\
\text{where} \quad K_m &= \{ (m-1)\Delta K + 1, \dots, m\Delta K\}.
\end{aligned}
\label{eq:Tfun}
\end{equation}

\noindent Moreover, given the initial patch size $p_0$, initial stride $s_0$, patch size change rate $\Delta p$, and stride change rate $\Delta s$, we define the patch size $p_t$ and stride $s_t$ at each time step $t$ as follows:
\begin{equation}
   p_t = \left\lfloor \frac{t-1}{\Delta K}  \right\rfloor \cdot \Delta p + p_0, \text{ } s_t= \left\lfloor \frac{t-1}{\Delta K} \right\rfloor \cdot \Delta s + s_0. 
\label{eq:pfun}
\end{equation}

\begin{algorithm}[t]
\scriptsize
\caption{P-PatchDiff Sampling}\label{alg:psample}
\begin{algorithmic}[1]
\Require low-light image $\mathbf{y}$, diffusion model $f_\theta$, estimator $g$, encoder $h$.
\State Sample $\mathbf{x}_t \sim \mathcal{N}(\mathbf{0}, \mathbf{I})$
\For{$t = T, \dots, 1$}
\State Obtain $t$ and $t_{\text{next}}$ by DDIM
\State Obtain $p_t$ and $s_t$ by \Cref{eq:pfun}
\State $\mathbf{F}_{ill} = h(g(\mathbf{y}_{p_t}))$
\State Generate dictionary of $D^p_s$ overlapping patch locations with size $p_t$ and stride $s_t$
\State $\bm{\hat\Omega}_t = 0$ and $\mathbf{M} = 0$
\For{$d = 1, \dots, D^p_s$}
\State Crop $\mathbf{x}_t^{(d)} = \mathbf{P}_d \odot \mathbf{x}_t$, $\mathbf{y}^{(d)} = \mathbf{P}_d \odot \mathbf{y}$
\State $\bm{\hat\Omega}_t = \bm{\hat\Omega}_t + \mathbf{P}_d \odot f_\theta(\mathbf{x}_t^{(d)}, \mathbf{y}^{(d)}, \mathbf{F}_{ill}, t)$
\State $\mathbf{M} = \mathbf{M} + \mathbf{P}_d$
\EndFor
\State $\bm{\hat\Omega}_t = \bm{\hat\Omega}_t \oslash \mathbf{M}$ \Comment{$\oslash$: element-wise division}
\State $\mathbf{x}_t \gets \sqrt{\bar{\alpha}_{t_{\text{next}}}} \left(\frac{\mathbf{x}_t - \sqrt{1-\bar{\alpha}_t}\cdot\bm{\hat\Omega}_t}{\sqrt{\bar{\alpha}_t}}\right) + \sqrt{1-\bar{\alpha}_{t_{\text{next}}}} \cdot \bm{\hat\Omega}_t$
\EndFor
\State \textbf{return} $\mathbf{x}_t$
\end{algorithmic}
\end{algorithm}

As outlined in Line 3 of \Cref{alg:ptrain}, the main distinction between our P-PatchDiff and standard PDMs is that $p$ in P-PatchDiff 
progressively increases over time steps. We follow \citep{weatherdiff} to learn a conditional diffusion model where we provide a low-light patch $\textbf{y}^{(i)}$ as input, such that the sampled output has high fidelity to the data distribution conditioned on $\textbf{y}^{(i)}$. Additionally, as outlined in Lines 4 and 6 of  \Cref{alg:psample}, unlike previous methods that use a fixed $s$, $\textbf{x}_t$ is cropped into overlapping patches with dynamic stride $s_t$ during each sampling time step. This formulation ensures balanced detail and efficiency for low-light enhancement. This stems from the fact that multi-level information can be introduced through a progressive increase in patch size, which gradually shifts from local to global views, rather than cropping multiple patch scales in a single denoising step. Additionally, using a dynamic $s_t$ prevents the generation of too many overlapping patches when $p_t$ is large.
Thus, our P-PatchDiff provides a GPU-friendly, scalable solution for low-light enhancement that integrates multiple receptive fields without sacrificing efficiency or quality, making it a highly adaptable framework for diverse image sizes and settings.

Importantly, the formulations in \Cref{eq:Tfun} and \Cref{eq:pfun} define a general patch diffusion model in which existing methods emerge as special cases. For example, 
WeatherDiff~\citep{weatherdiff} is a special case of our P-PatchDiff when both $\Delta p$ and $\Delta s$ are set to $0$.

Finally, following SR3~\citep{SR3}, we train P-PatchDiff in a conditional DDPM manner. Specifically, P-PatchDiff predicts the noise $\bm{\epsilon}_t$ added to $\mathbf{x}_0^{(i)}$, where $\bm{\epsilon}_t\sim\mathcal{N}(\mathbf{0},\mathbf{I})$ is the noise sampled at time step $t$, and $\mathbf{x}_0^{(i)}$ is one patch of the ground-truth normal-light image. The full loss function is given in Line 6 of \Cref{alg:ptrain}, where $(\mathbf{x}_0, \mathbf{y})$ denotes the paired normal- and low-light images.

\subsection{Multi-patch alignment}
\label{sec:multipatch}
As discussed earlier,  without a global brightness reference, PDMs struggle to balance local and global brightness, particularly when processing high-resolution images, as shown in \Cref{fig:comparehigh}. Therefore, it is important to carefully manage outputs from the dynamic $p_t$ during the denoising process.
Moreover, as demonstrated by~\citep{pydiff}, while diffusion models can effectively estimate noise at each time step even when input sizes vary, they remain sensitive to colour shifts.  We therefore introduce a global brightness proxy to serve as an unchanging reference for all patches across all denoising steps.

\begin{table*}[t]
\fontsize{6}{7.2}
\selectfont
\caption{Quantitative comparisons on the LOL-v1, LOL-v2-Real and LOL-v2-Syn datasets in terms of PSNR and SSIM. \textbf{Bold} and \underline{underline} denote the best and second-best results. The top half and bottom half of the table show regression and generative models, respectively. $\diamond$ denotes patch diffusion models.}
\centering
    \begin{adjustbox}{width=\linewidth}
    \begin{tabular}{c|c|c|c|c|c|c}
    \toprule
    \multirow{2}{*}{Methods} & \multicolumn{2}{c}{LOL-v1 ($400 \times 600$)} & \multicolumn{2}{|c|}{LOL-v2-Real ($400 \times 600$)} &  \multicolumn{2}{c}{LOL-v2-Syn ($512 \times 512$)} \\
    \cmidrule{2-7}
  & PSNR $\uparrow$  & SSIM $\uparrow$  & PSNR $\uparrow$  & SSIM $\uparrow$ & PSNR $\uparrow$ & SSIM $\uparrow$  \\
    \midrule
     RetinexNet~\citep{RetinexNet} & 16.77 & 0.462 & 17.72 & 0.652 & 16.55 & 0.652 \\
     KinD~\citep{KinD} & 20.87 & 0.799 & 17.54 & 0.669 & 18.96 & 0.801 \\
     Zero-DCE~\citep{ZeroDCE} & 14.86 & 0.562 & 18.06 & 0.580 & 17.76 & 0.814 \\
    RUAS~\citep{RUAS} & 16.41 & 0.503 & 15.35 & 0.495 & 13.40 & 0.640 \\
     SCI~\citep{SCI} & 14.78 & 0.525 & 17.30 & 0.540 & 15.43 & 0.744  \\
     URetinexNet~\citep{URetinexNet} & 19.84 & 0.824 & 21.09 & \underline{0.858} & 18.27 & 0.518 \\
    SNR~\citep{SNR} & \underline{24.61} & \underline{0.842} & 21.48 & 0.849 & 24.14 & 0.927  \\
    SMG~\citep{smg} & 23.68 & 0.826 & \textbf{24.62} & \textbf{0.867} & \underline{25.62} & \underline{0.905}  \\ 
    PairLIE~\citep{PairLIE} & 19.51 & 0.736 & 19.88 & 0.773 & 19.07 & 0.794   \\
    Retinexformer~\citep{Retinexformer} & \textbf{25.15} & \textbf{0.843}  & \underline{22.79} & 0.839 & \textbf{25.67} & \textbf{0.928} \\
    Zero-IG~\citep{zeroig} & 22.17 & 0.772 & 18.13 &	0.740 & 15.77 & 0.752 \\
    \midrule
    \midrule      
     EnlightenGAN~\citep{EnlightenGAN} & 17.61 & 0.653 & 18.68 & 0.678  & 16.57 &	0.772 \\ 
    PyDiff~\citep{pydiff} & 27.09 & 0.880 & \underline{26.98} & \underline{0.882} & \underline{25.28} & \underline{0.913}  \\
    DiffLL~\citep{wavediff} & 26.34 & 0.845 & 19.89 & 0.806  & 22.67 & 0.869 \\
    WeatherDiff~\citep{weatherdiff} $\diamond$ &  21.44 &  0.837 & 22.01 & 0.838  & 22.33 & 0.909 \\
    CLEDiff~\citep{clediff} & 24.92 & 0.880 & 20.94 &	0.794  & 20.59 & 0.773 \\
    MDMS~\citep{MDMS} $\diamond$ & \underline{27.12} & \textbf{0.882}  & 26.87 & 0.871  & 19.83 & 0.825 \\
    QuadPrior~\citep{quadpior} & 20.31 & 0.808  & 20.59 &	0.808  & 16.10 & 0.753 \\
    LightenDiff~\citep{lightendiff} & 20.45 & 0.803 & 22.93 &	0.853 & 21.57 &	0.866  \\ 
     P-PatchDiff (Ours) $\diamond$ & \textbf{27.18} & \underline{0.880} & \textbf{28.04} & \textbf{0.887} & \textbf{28.15} & \textbf{0.941} \\
    \bottomrule
    \end{tabular}
    \end{adjustbox}
    \label{tab:comparelol1}

\end{table*}

Specifically, as shown in \Cref{fig:overall} (c), following~\citep{Retinexformer,GSAD}, we compute the ground-truth brightness 
proxy by taking the mean value along the channel dimension of the normal-light
image $\mathbf{x}_0$. Unlike other methods that model illumination using physical theory~\citep{RetinexNet,guo2025exploring}, we adopt this channel-wise mean as a coarse but lightweight prior, favouring simplicity and ease of deployment. Despite its simplicity, we find that it effectively handles local enhancement while maintaining  global coherence. One reason is that we perform alignment at the feature level, which provides high-dimensional and robust information of low-light images, rather than the pixel level~\citep{pydiff}. Moreover, we learn a lightweight U-Net~\citep{unet} $g$ to predict $\textbf{x}_{ill}$  from $\mathbf{y}_{p_t}$, which is downsampled to size $p_t \times p_t$ from the low-light full-size image:
\begin{equation}
   \mathcal{L}_{ill} = |\text{mean}({\textbf{x}}_0) -  \textbf{x}_{ill}|, \text{where} \quad \textbf{x}_{ill}= g(\textbf{y}_{pt}).
\end{equation}

\noindent Here $\mathbf{x}_0$ is also downsampled to $p_t \times p_t$ for supervision. By learning from real-world normal-light data, $g$ captures realistic statistics of normal-light images, thereby improving adaptability during inference. This mitigates the difficulty of relying on a handcrafted illumination prior when enhancing unseen images~\citep{clediff}. 
Importantly, $g$ is only run once per denoising step to produce $\mathbf{x}_{{ill}}$ that is reused for all patches, incurring minimal time cost.

\begin{table*}[t]
\fontsize{6}{7.2}
\selectfont
   \caption{Quantitative comparisons on the LOL-v1, LOL-v2-Real and LOL-v2-Syn datasets in terms of LPIPS and FID. \textbf{Bold} and \underline{underline} denote the best and second-best results. The top half and bottom half of the table show regression and generative models, respectively.  $\diamond$ denotes patch diffusion models.}
    \centering
    \begin{adjustbox}{width=\linewidth}
    \begin{tabular}{c|c|c|c|c|c|c}
    \toprule
    \multirow{2}{*}{Methods} & \multicolumn{2}{c}{LOL-v1 ($400 \times 600$)} & \multicolumn{2}{|c|}{LOL-v2-Real ($400 \times 600$)} &  \multicolumn{2}{c}{LOL-v2-Syn ($512 \times 512$)} \\
    \cmidrule{2-7}
   & LPIPS $\downarrow$ & FID$\downarrow$  & LPIPS $\downarrow$ & FID $\downarrow$ & LPIPS $\downarrow$ & FID $\downarrow$ \\
    \midrule
     RetinexNet~\citep{RetinexNet}  & 0.417 & 126.27 & 0.436 & 133.91  & 0.379 & 98.84 \\
     KinD~\citep{KinD} & 0.207 & 104.63 & 0.375 & 137.35  & 0.262 & 89.16\\
     Zero-DCE~\citep{ZeroDCE}  &  0.335 & 101.24  & 0.313 & 91.94 & 0.168 & 49.24 \\
    RUAS~\citep{RUAS}  & 0.364 & 101.97 & 0.395 & 94.16 & 0.364 & 123.78\\
     SCI~\citep{SCI}  & 0.366 & 78.60  & 0.345 & 67.62 	& 0.233 & 61.20  \\
     URetinexNet~\citep{URetinexNet}  & 0.237 & \textbf{52.38}  & 0.208 & \textbf{49.84}  & 0.419 & 66.87 \\
    SNR~\citep{SNR}  & 0.233 & \underline{55.12} & 0.237 & \underline{54.53}  & 0.056 & \textbf{19.95} \\
    SMG~\citep{smg} & \textbf{0.118} & 58.85 & \textbf{0.148} & 78.58  & \textbf{0.053} & 23.21 \\ 
    PairLIE~\citep{PairLIE}  & 0.248 & 103.64 & 0.239 & 98.27 & 0.232 & 86.41   \\
    Retinexformer~\citep{Retinexformer} & \underline{0.131} & 71.15  & \underline{0.171} & 62.44  & \underline{0.059} & \underline{22.78} \\
    Zero-IG~\citep{zeroig} & 0.199 & 83.94  & 0.248 &	77.38  &0.259 &	78.49\\
    \midrule
    \midrule      
     EnlightenGAN~\citep{EnlightenGAN}& 0.372 & 94.70  & 0.364 & 84.04  &	0.212 & 74.35 \\ 
    PyDiff~\citep{pydiff} & 0.100 & 48.38  & 0.154 & 77.25  & 0.095 & 36.53 \\
    DiffLL~\citep{wavediff} & 0.217 & \underline{48.11} & 0.202 & 81.47 & 0.154 & 63.49 \\
    WeatherDiff~\citep{weatherdiff} $\diamond$  &  0.159 & 78.48  & 0.204 & 89.82 & \underline{0.092} & \underline{35.44} \\
    CLEDiff~\citep{clediff}  & 0.160 & 74.29  &	0.178 &	71.25  &	0.273	& 101.40 \\
    MDMS~\citep{MDMS} $\diamond$ &  \textbf{0.078} & \textbf{36.29} & \textbf{0.101} & \underline{69.12}  & 0.225 & 81.74 \\
    QuadPrior~\citep{quadpior}  & 0.202 & 79.07  & 0.202 &	70.90 & 0.250 & 76.76 \\
    LightenDiff~\citep{lightendiff} & 0.192 & 85.37  &	0.167 &	79.04  &	0.155	& 56.97 \\ 
     P-PatchDiff (Ours) $\diamond$  & \underline{0.096} & 52.35  & \underline{0.135} &  \textbf{66.02} & \textbf{0.054} & \textbf{23.90} \\
    \bottomrule
    \end{tabular}
    \end{adjustbox}
    \label{tab:comparelol2}
\end{table*}

Since $\mathbf{x}_{{ill}}$ mixes multiple attributes such as illumination, object colour, and exposure, it is challenging to directly use it for alignment. We thus employ a stack of CNNs $h$ to obtain $\textbf{F}_{ill}$ from $\mathbf{x}_{{ill}}$. Inspired by the conditional diffusion model, we provide $\textbf{F}_{ill}$ into $f_\theta$ as conditional input.
As shown in \Cref{fig:overall} (d), we integrate $\mathbf{F}_{ill}$ into each encoder layer of $f_\theta$
through a learnable multi-patch alignment layer (PALayer) to adaptively adjust features. Given the encoder feature
$\mathbf{F}_l$ and $\mathbf{F}_{ill}$, PALayer predicts channel-wise
scaling and bias vectors $(w_l, b_l)$ via convolutional layers, and
normalises $\mathbf{F}_l$ as:
\begin{equation}
   \hat{\mathbf{F}}_l = w_l \odot \text{GroupNorm}(\mathbf{F}_l) + b_l.
\end{equation}
 This operation aligns the statistics of each patch's features to the global intensity, mitigating colour and brightness inconsistencies across
different $p_t$ values. 

The overall objective function of P-PatchDiff is:
\begin{equation}
    \mathcal{L}_{total} = \mathcal{L}_{diff} + \mathcal{L}_{ill}, 
\end{equation}
\noindent where $\mathcal{L}_{diff}$ refers to Line 6 in \Cref{alg:ptrain}.

\begin{table*} 
\fontsize{6}{7.2}
\selectfont
    \caption{Quantitative results on the LSRW and UHD-LL datasets. Following~\citep{wavediff}, images are downsampled for methods that run out of memory. $\dagger$ denotes methods that support 4K.}
    \label{tab:comparelsrw}
    \centering
    \begin{adjustbox}{width=\linewidth}
    \begin{tabular}{c|c|c|c|c|c|c|c|c}
    \toprule
        \multirow{2}{*}{Methods} & \multicolumn{4}{c|}{LSRW ($960 \times 720$)} & \multicolumn{4}{c}{UHD-LL (4K)}  \\
        \cmidrule{2-9}
         & PSNR  $\uparrow$ & SSIM $\uparrow$  & LPIPS $\downarrow$ & FID $\downarrow$ & PSNR $\uparrow$ & SSIM $\uparrow$ & LPIPS $\downarrow$ & FID $\downarrow$ \\
        \midrule
        SNR~\citep{SNR} & \underline{17.30} & 0.512 & 0.452 &	94.19 &16.98 &	0.797 &	0.245 &	69.90 \\
        PyDiff~\citep{pydiff} & 16.88 & 0.503 & 0.328 & 73.02 & 19.64 & 0.844 & 0.180 & 46.35 \\
        Zero-IG~\citep{zeroig} $\dagger$ & 15.90 & 0.485 &	0.364 & 74.89 & 13.79 &	0.718 &	0.305 & 80.73 \\
        DiffLL~\citep{wavediff} $\dagger$ & 17.04 & 0.390 &	0.430	& 80.66 & \underline{21.49} & 0.721 &	0.336 &	63.30 \\
        CLEDiff~\citep{clediff}& 13.97 & 0.425 &	0.502 &	100.31 & 11.97 & 0.641 &	0.278 &	65.76 \\
        GSAD~\citep{GSAD} & 16.99 & 0.494 & \textbf{0.290} & \underline{65.18} & 21.27 & \underline{0.845} & 0.171 & \underline{37.60} \\
        WeatherDiff~\citep{weatherdiff} $\diamond$ $\dagger$ &16.37 &	\underline{0.514} & 0.310 & 77.18 & 17.22	&0.820 & 0.202 &  65.12 \\
        MDMS~\citep{MDMS} $\diamond$ $\dagger$ & 15.81 &	0.504 &	0.319 &	68.16 & 17.93 & 0.832 & \underline{0.177} & 44.80 \\
        P-PatchDiff (Ours) $\diamond$ $\dagger$ & \textbf{17.61} & \textbf{0.520} &	\underline{0.298} & \textbf{64.63} & \textbf{21.79} &	\textbf{0.857} &	\textbf{0.150} &	\textbf{35.55} \\
        \bottomrule
    \end{tabular}
    \end{adjustbox}
\end{table*}

\begin{table*}[t]
\fontsize{6}{7.2}
\selectfont
\caption{Quantitative comparisons on the DICM, MEF, LIME, NPE and VV datasets. \textbf{Bold} and \underline{underline} denote the best and second-best results. NI, BR, PI denote the NIQE, BRISQUE and PI metrics, respectively.  $\diamond$ denotes patch diffusion models.}
    \centering
    \begin{adjustbox}{width=\linewidth}
    \begin{tabular}{c|c|c|c|c|c|c|c|c|c}
    \toprule
    \multirow{2}{*}{Methods} & \multicolumn{3}{c}{DICM} & \multicolumn{3}{|c|}{MEF} &  \multicolumn{3}{c}{LIME}  \\
    \cmidrule{2-10}
    & NI $\downarrow$ & BR $\downarrow$ & PI $\downarrow$ & NI $\downarrow$ & BR $\downarrow$ & PI $\downarrow$ & NI $\downarrow$ & BR $\downarrow$ & PI $\downarrow$  \\
    \midrule
    SNR~\citep{SNR} & \textbf{3.43} &	34.48 &	4.11 &	\underline{3.52} & 30.57 &	3.55 &	4.77 &	38.69 &	4.64  \\
     SMG~\citep{smg} & 4.46 &	\textbf{24.85} & 3.71 & 5.09& \underline{21.27} & 3.77 &	4.96 &	30.93 & 4.11  \\
     PyDiff~\citep{pydiff} & 3.57 & 30.37 &	3.67 &		\textbf{3.49} &	30.24 &	3.64 &	4.99 &	38.57 &	5.34  \\
     CLEDiff~\citep{clediff} & 5.61 &	42.46 &	4.77 &		4.77 &	\textbf{15.56} &	3.68 &		7.74 &	38.26 &	6.96  \\
     QuadPrior~\citep{quadpior} & 4.07 &	25.58 & \underline{3.65} &	3.65 &	22.49 &	\underline{3.13} &	4.59 &	\underline{28.02} &	3.93  \\
     WeatherDiff~\citep{weatherdiff} $\diamond$ & 3.77 & 30.38 & 4.13 & 3.75 & 30.48 & 3.31 & 5.31 & 28.09 & 4.42 \\
     MDMS~\citep{MDMS} $\diamond$ & 3.85	& \underline{25.19} &	\textbf{3.45} &		4.11&	23.45 &	\textbf{3.12} &	\textbf{4.19} &	29.56 &	\underline{3.78}  \\
      P-PatchDiff (Ours) $\diamond$ & \underline{3.53} &	30.40 &	3.74 &		3.54 &	28.42 &	3.52 &		\underline{4.59} &	\textbf{27.53} &	\textbf{3.68}  \\
    \bottomrule
    \end{tabular}
    \end{adjustbox}
    \label{tab:comparewogt}

    \begin{adjustbox}{width=\linewidth}
    \begin{tabular}{c|c|c|c|c|c|c|c|c|c}
    \toprule
    \multirow{2}{*}{Methods} & \multicolumn{3}{c}{NPE} & \multicolumn{3}{|c}{VV} & \multicolumn{3}{|c}{Average (5 datasets)} \\
    \cmidrule{2-10}
    & NI $\downarrow$ & BR $\downarrow$ & PI $\downarrow$ & NI $\downarrow$ & BR $\downarrow$ & PI $\downarrow$ & NI $\downarrow$ & BR $\downarrow$ & PI $\downarrow$  \\
    \midrule
    SNR~\citep{SNR}  &	\underline{3.51} & 26.72 & 3.39 &		5.31 & 56.94 & 7.68 &	4.11 & 37.48 & 4.67 \\
     SMG~\citep{smg}   & 4.25 & \underline{25.85}	& 3.55 &		5.92	& 54.74 &7.09	& 4.94 & 31.53 & 4.44 \\
     PyDiff~\citep{pydiff}  &	3.61 &	26.19 &	3.43 &		\underline{4.12}	& 37.61 &	5.73 &	\underline{3.95}	&32.60 &	4.36 \\
     CLEDiff~\citep{clediff}  &		4.93 &	\textbf{24.03} & 3.33 &		7.40 &	42.25 &	6.62 &		6.09 & 32.51 &	5.07 \\
     QuadPrior~\citep{quadpior}   &		3.52 &	28.37 &	\textbf{3.12} &	5.00 &	47.70 &	6.84 & 4.16 &	\underline{30.43} &	4.13 \\
     WeatherDiff~\citep{weatherdiff} $\diamond$   & 3.67  & 30.26  & 3.87 & 3.47 & 38.07 & 4.65 & 3.99 &31.46 &	\underline{4.08}\\
     MDMS~\citep{MDMS} $\diamond$  & 4.13 & 28.56&  \underline{3.15} &	5.53 &	48.28 & 7.46 &		4.36 &	31.01 &	4.19 \\
      P-PatchDiff (Ours) $\diamond$   &		\textbf{3.41} &	30.01 & 3.26 &		\textbf{2.62} &	\textbf{28.97} &	\textbf{3.82} &		\textbf{3.54} &	\textbf{29.07} &	\textbf{3.61} \\
    \bottomrule
    \end{tabular}
    \end{adjustbox}
    
\end{table*}

\section{Experiments}
To validate the effectiveness of the proposed P-PatchDiff, we conduct experiments under two different settings: (1) In-domain evaluation, where models are trained and evaluated on the official training and test sets for low-light image enhancement; and (2) Cross-domain generalisation, where models are trained on one dataset and evaluated on other low-light image enhancement datasets.

\subsection{Implementation details}
\textbf{Dataset. } For low-light image enhancement, we evaluate P-PatchDiff on 9 real-world and 1 synthetic widely used public datasets with \textbf{varying image resolutions}.

For in-domain evaluation, we train and test P-PatchDiff on \textbf{LOL-v1}~\citep{RetinexNet} and \textbf{LOL-v2-Real}~\citep{LOLv2}, both with a size of 400 × 600 and \textbf{dynamic ISO}, as well as \textbf{LOL-v2-Syn}~\citep{LOLv2}, with a size of 512 × 512. These datasets include diverse indoor and outdoor city scenes, with training/testing splits of 485:15, 689:100, and 900:100, respectively. 

For cross-domain generalisation, we test our model (trained on LOL-v1) on \textbf{LSRW}~\citep{lsrw}, which contains 50 images at 960 × 720 captured by Nikon cameras and a mobile phone with \textbf{lower ISO ([50, 100])}, and \textbf{UHD-LL}~\citep{uhdlol}, which consists of 150 \textbf{4K} images captured by Sony cameras with \textbf{higher ISO ([100, 800])}. Furthermore, we evaluate the same model on five real-world datasets without ground truth, covering world-wide low-light scenarios and more complex resolutions: \textbf{DICM}~\citep{DICM} at 480 × 640, \textbf{MEF}~\citep{MEF} at 512 × 340, \textbf{LIME}~\citep{LIME} at 2000 × 1500, \textbf{NPE}~\citep{NPE} at 725 × 750, and \textbf{VV}~\citep{VV} at 2304 × 1728.

\textbf{Metrics. }  We use PSNR, SSIM, LPIPS and FID for datasets with ground truth, while NIQE~\citep{NIQE}, BRISQUE~\citep{BRI}, and PI~\citep{PI} are used for datasets without ground truth.

\textbf{Implementation. } Our method is built upon WeatherDiff~\citep{weatherdiff} and trained on an NVIDIA A100 GPU with 80G memory for 500k iterations. We use the Adam optimiser with an initial learning rate of $1 \times 10^{-4}$ and halve the learning rate every 50k iterations. The batch size is set to 16, with training timesteps $T=1000$. \textit{We use DDIM~\citep{ddim} for sampling with only 5 steps}. 

\textbf{Hyperparameters in progressive strategy. } 
We follow existing PDMs~\citep{weatherdiff,MDMS} by setting $p_0=64$ and $s_0=16$, ensuring that our method starts from the same initial patch size and stride during sampling for a fair comparison. Furthermore, we adopt $\Delta p = 32$, consistent with MDMS, so that both methods use the same set of multi-scale patches. These settings allow us to isolate the effect of our progressive strategy by varying only $n$, while keeping the patch configuration identical to other PDMs. Based on the analysis in \Cref{fig:psstudy}, we set $\Delta s = 16$ to ensure that $s_t$ remains less than half of $p_t$ throughout the process. We use $n=5$ for optimal performance, with the corresponding ablation results presented in \Cref{tab:abpatch}.

\subsection{Comparisons with existing methods}
We compare P-PatchDiff with state-of-the-art regression and generative models. \textbf{For regression-based methods}, we select CNNs (RetinexNet~\citep{RetinexNet}, KinD~\citep{KinD} and URetinexNet~\citep{URetinexNet}), Transformers (SNR~\citep{SNR},  SMG~\citep{smg} and Retinexformer~\citep{Retinexformer}) and zero-reference methods (Zero-DCE~\citep{ZeroDCE},  RUAS~\citep{RUAS}, SCI~\citep{SCI}, PairLIE~\citep{PairLIE} and Zero-IG~\citep{zeroig}). \textbf{For generative-based models}, we select EnlightenGAN~\citep{EnlightenGAN} and DDPMs: Directly generating normal-light images (PyDiff~\citep{pydiff}, GSAD~\citep{GSAD}, CLEDiff~\citep{clediff} and  QuadPrior~\citep{quadpior}), generating components (DiffLL~\citep{wavediff} and  LightenDiff~\citep{lightendiff}) and PDMs (WeatherDiff~\citep{weatherdiff} and 
MDMS~\citep{MDMS}) \textit{denoted by $\diamond$ in each table}.

\textbf{In-domain evaluation. } As shown in \Cref{tab:comparelol1,tab:comparelol2}, P-PatchDiff achieves competitive results across three popular low-resolution benchmarks with multi-exposure images for in-domain testing. The best-performing PDM, MDMS, incorporates multi-scale patches together with a Fourier Transform, which can be particularly beneficial under supervised settings. In contrast, our method adopts a standard U-Net backbone, which prioritises flexibility and general applicability, but may limit performance in certain fully supervised benchmarks. Moreover, MDMS requires significantly higher computational costs (600× slower than ours), making it impractical for real-world use. DiffLL performs well on $400 \times 600$ datasets but fails on 4K resolutions in terms of SSIM/LPIPS (0.1 lower SSIM and 0.16 higher LPIPS than ours). This is because they rely on transformations to resize images for efficiency, which results in the loss of fine details.

\textbf{Cross-domain generalisation. } As shown in \Cref{tab:comparelsrw}, we also show competitive performance for cross-dataset validation on LSRW and UHD-LL, which include diverse ISO and 4K resolution images. In comparison, although CLEDiff incorporates illumination information into the denoising process, it can only operate effectively within the training domain and fails to generalise to LSRW and UHD-LL. QuadPrior proposes a general illumination prior to improve generalisability, but it relies on downsampled single-scale images, which harms performance. Even though our estimator $g$ shares a similar concept with these diffusion models by injecting priors, we employ it more effectively, using $\mathbf{F}_{ill}$ to align multi-level features through the multi-patch alignment layer, thereby improving consistency across scales.

\begin{table}[t]
\fontsize{6}{7.2}
\selectfont
    \caption{Detailed training statistics on the LOL-v2-real dataset, including training time (T.t),  memory usage (M), and parameters (P). }  
    \label{tab:abcost}
    \centering
    \begin{tabular}{c|c|c|c|c}
    \toprule
        Methods & PSNR $\uparrow$ & T.t (h) $\downarrow$ & M (G)  $\downarrow$ & P (M) $\downarrow$  \\
        \midrule
        PyDiff & 26.98 & 98  & 20 & 97 \\
        CLEDiff & 20.94 & 72  & 32 & 85 \\ 
        QuadPrior & 20.59 & - & 40 & 1313 \\
        LightenDiff & 22.93 & 69  & 11 & \textbf{28} \\ 
       WeatherDiff $\diamond$ & 22.01 & 106  & 8 & 61 \\
       MDMS $\diamond$  &   26.87 & 145   & 23 & 213 \\
       Ours $\diamond$  & \textbf{28.04} & \underline{55} & \underline{8} & 80 \\
       Ours-S $\diamond$  & \underline{27.64} & \textbf{17} & \textbf{6} & \underline{31} \\
       \bottomrule
    \end{tabular}
\end{table}

As illustrated in \Cref{tab:comparewogt}, we further conduct experiments on five complex and uncommon-resolution datasets. We note that some methods, such as SMG, CLEDiff, and QuadPrior, show better results than ours in some cases. This arises because these methods include pretrained knowledge or handcrafted priors that align closely with certain target data distributions. However, such designs might not generalise consistently across diverse domains. In contrast, our method emphasises progressive multi-scale patch modelling and global coherence to improve robustness under varying illumination conditions, which leads to competitive and often superior overall performance. Additionally, their low performance on NPE and VV datasets indicates ineffectiveness in handling high-resolution images.

\begin{figure*}[t]
\centering
\includegraphics[width=\linewidth]{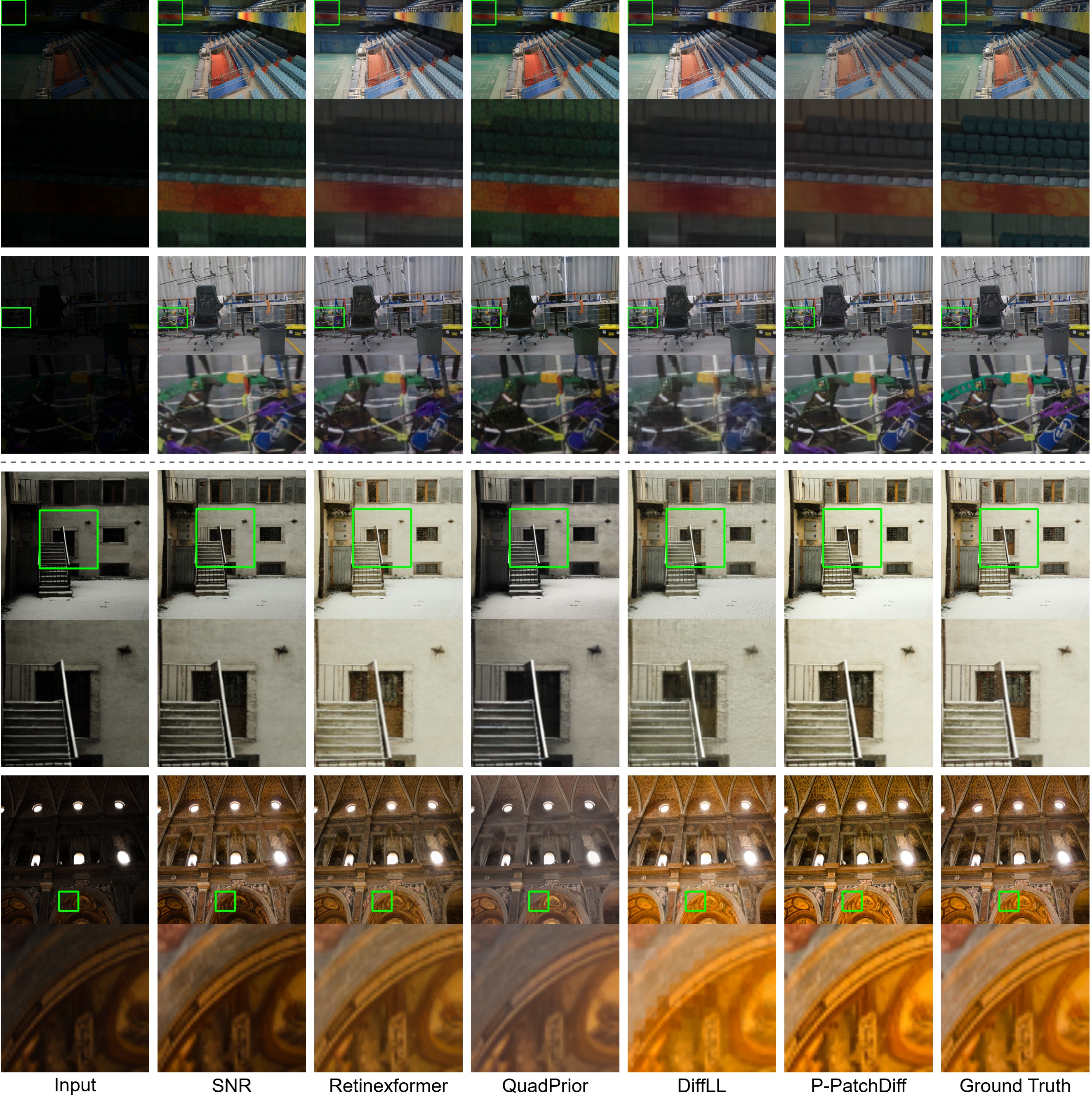}
\caption{Qualitative comparison on the LOL-v2-Real (above the dashed line) and LOL-v2-Syn (below the dashed line).}
\label{fig:comparelol}
\end{figure*}

\begin{figure*}[t]
\centering
\includegraphics[width=\linewidth]{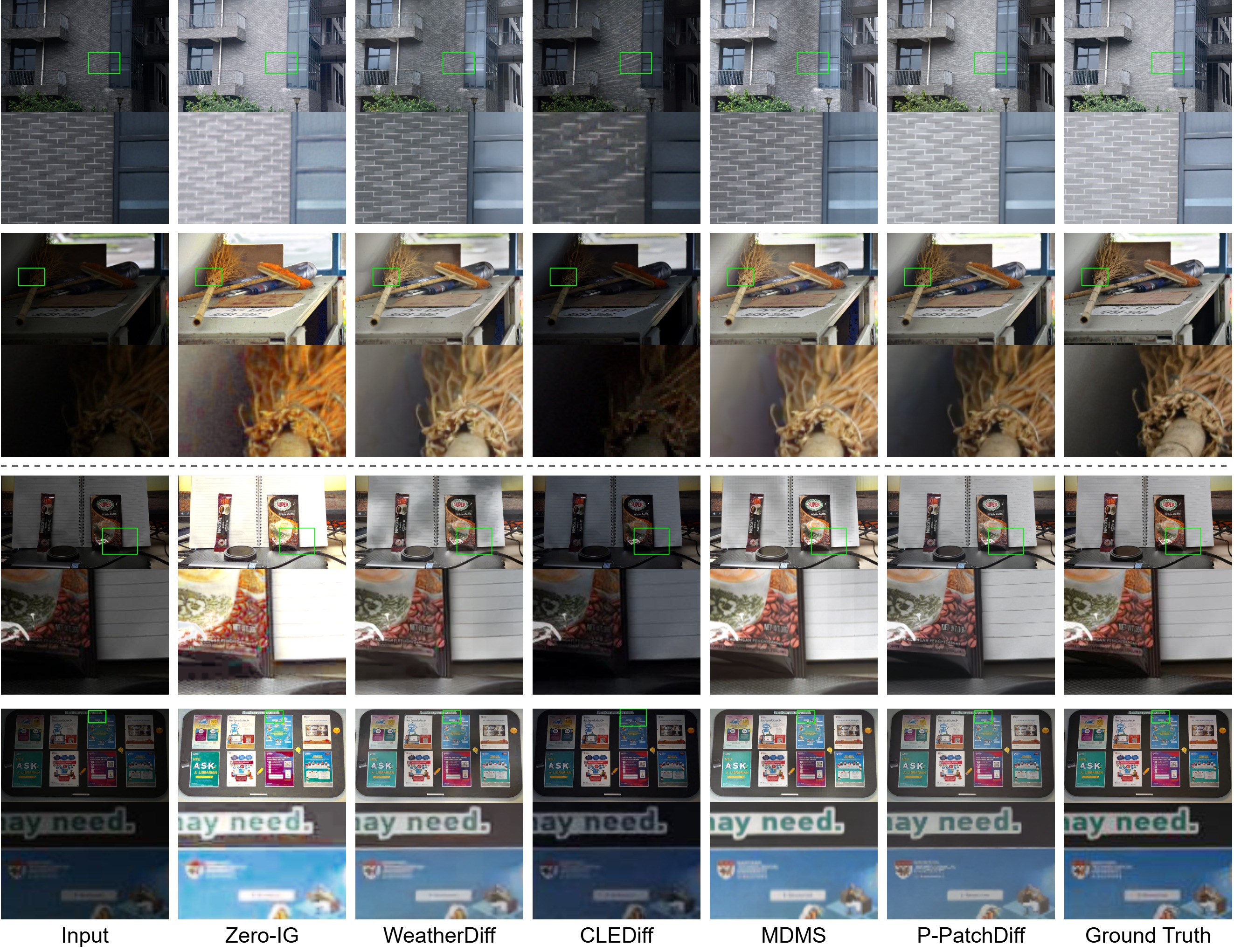}
\caption{Qualitative comparison on the LSRW (above the dashed line) and UHD-LL (below the dashed line).}
\label{fig:comparehigh}
\end{figure*}

\begin{figure*}[t]
\centering
\includegraphics[width=1\linewidth]{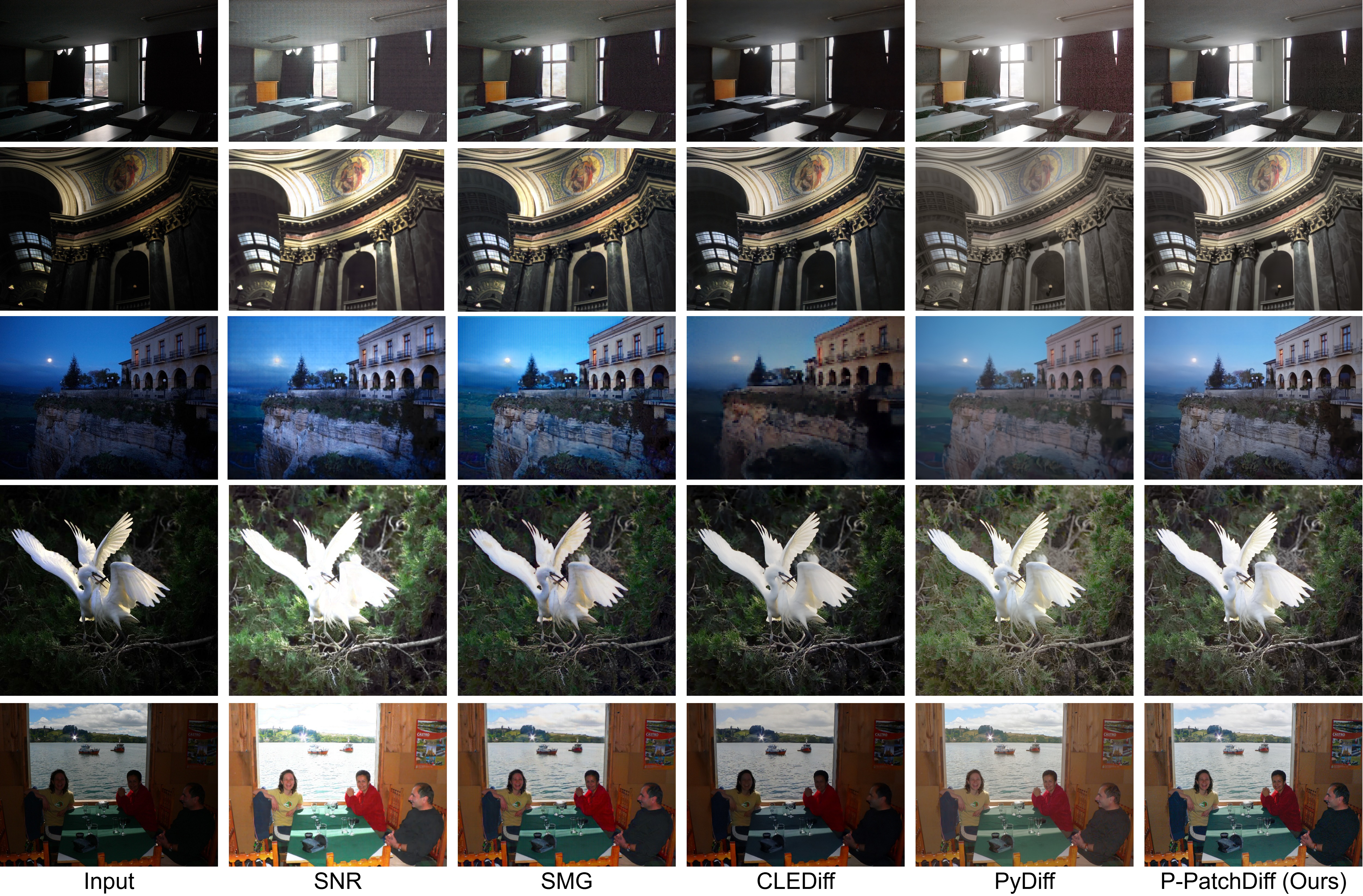}
\caption{Visual results on the DICM, MEF, LIME, NPE and VV datasets (From top to bottom).}
\label{fig:comparewogt}
\end{figure*}

\textbf{Qualitative performance. } In \Cref{fig:comparelol,fig:comparehigh}, P-PatchDiff produces better colour restoration than these single-scale methods and achieves performance comparable to multi-scale methods (e.g., MDMS) while maintaining much faster processing.  
In \Cref{fig:comparewogt}, it is evident that CLEDiff tends to generate undersaturated images. Moreover, PyDiff renders the two curtains in inconsistent colours. In the third row, it also generates an overly smooth image. In contrast, P-PatchDiff demonstrates robustness in challenging scenarios, consistently restoring natural colours and preserving fine structures across diverse conditions and resolutions.

\begin{table}[t]
\addtolength{\tabcolsep}{-.8ex}
\fontsize{6}{7.2}
\selectfont
    \caption{Detailed sampling statistics, including memory usage (M) and sampling time (S.t) with different input sizes. OOM indicates out-of-memory error. We use an NVIDIA A100 GPU with 80G RAM.}  
    \label{tab:abvaryingcost}
    \centering
     \begin{tabular}{c|c|c|c|c|c|c}
    \toprule
        Input sizes & \multicolumn{2}{c}{$400 \times 600$} & \multicolumn{2}{|c|}{1080p} & \multicolumn{2}{c}{4K}  \\
        \midrule
        Method & M (G)  & S.t (s) & M (G)  & S.t (s) & M (G)  & S.t (s)  \\
        \midrule
        PyDiff & 6.5 & 2.1 & 16.1 & 4.3 & OOM & OOM \\
        CLEDiff & 5.8 & 32.0 & 62.6 & 67.5 & OOM & OOM \\ 
        QuadPrior & 10.2 & 4.3 & OOM & OOM & OOM & OOM \\
        LightenDiff & \textbf{2.6} & \textbf{0.3} & 9.1 & \textbf{0.9} & 19.9 & \textbf{4.5}  \\ 
       WeatherDiff $\diamond$  & 6.4 & 15.6 & 8.7 & 125.8 & 11.4 & 554.7 \\
       MDMS $\diamond$  &  16.2 & 1437.0 & 19.3 & 4597.0 & 51.1 & 8232.1  \\
       Ours $\diamond$  &  5.7 & 2.5 & \underline{6.3} & 20.0 & \underline{8.8} & 93.0 \\
          Ours-S $\diamond$  &  \underline{4.4} & \underline{0.5} & \textbf{5.3} & \underline{3.6} & \textbf{7.8} & \underline{15.0} \\
       \bottomrule
    \end{tabular}
\end{table}

\textbf{Computational costs. } As shown in \Cref{tab:abcost}, we compare diffusion methods in terms of training costs. Our progressive patchifying strategy achieves the lowest training time and memory consumption. This is because the patch size is dynamically adjusted to avoid consistently using large patches, and the estimator $g$ is designed to have minimal impact on sampling time. In \Cref{tab:abvaryingcost}, we evaluate P-PatchDiff's sampling costs across multiple resolutions, including 400 × 600, 1080p, and 4K. P-PatchDiff exhibits the lowest memory usage for 4K images, with only a gradual increase in sampling costs as resolution grows. Additionally, our costs are even less than WeatherDiff, which uses small fixed patches. This is because we improve efficiency by improving PDMs through progressive patches, which effectively reduce the number of cropped patches.

While LightenDiff reduces sampling time by transforming images to a smaller size, it still requires substantial memory and suffers from large performance drops across datasets. Notably, on LOL-v2-Real, its PSNR is over 5 dB lower than P-PatchDiff, indicating that it cannot maintain acceptable image quality. In contrast, we adopt the same U-Net backbone as~\citep{weatherdiff} but employ a progressive patch strategy, reducing computational cost while operating directly in the original image domain. For a fair comparison with other lightweight methods, we use the same channel multiplier as LightenDiff to obtain a model of similar size, denoted as Ours-S. This lightweight P-PatchDiff variant maintains performance close to the full model, while requiring significantly less memory, and still achieves higher PSNR gains compared to LightenDiff, alongside competitive inference times.

\begin{table*} 
\fontsize{6}{7.2}
\selectfont
    \caption{Comparison of different patch sampling methods.  S.t represents Sampling Time.} 
    \label{tab:abpatch}
    \centering
    \begin{adjustbox}{width=\linewidth}
    \begin{tabular}{c|c|c|c|c|c|c|c|c|c|c|c|c}
    \midrule
        \multirow{2}{*}{Methods} & \multicolumn{3}{c}{Fixed patch} &  \multicolumn{3}{|c|}{Multi-scale patch} &  \multicolumn{3}{c|}{Ours w/o $g$} &  \multicolumn{3}{c}{Ours}  \\
        \cmidrule{2-13}
        & PSNR & SSIM & S.t (s) & PSNR & SSIM & S.t (s) & PSNR & SSIM & S.t (s)  & PSNR & SSIM & S.t (s) \\
        \midrule
        $n=1$ & 25.35 & 0.864 & 3.60 & 25.35 & 0.864 & 3.60  & 25.35 & 0.864 & 3.60 & 27.06 & 0.874 & 3.61 \\
        $n=2$ & 25.82  & 0.873 & 12.82 & 25.81 & 0.871 & 18.24 & 26.17  & 0.875 & 2.72  & 27.50 & 0.875 & 3.24 \\
        $n=3$ & 25.31  & 0.864 & 17.84 & 26.00 & 0.874 & 35.50 & 26.64  & 0.882 & 2.39  & 27.57 & 0.875 & 2.59 \\
        $n=4$ & 26.69  & 0.873 & 23.18 & 26.21 & 0.877 & 57.56 & 26.81  & 0.884 & 2.39  & 27.94 & 0.885 & 2.59 \\
        $n=5$ & 26.54  & 0.880 & 28.48 & 26.46 & 0.879 & 85.12 & 27.04 & 0.885 & 2.24 & 28.04 & 0.887 & 2.45 \\
    \midrule
    \end{tabular}
    \end{adjustbox}
\end{table*}

\begin{table} 
\addtolength{\tabcolsep}{-.7ex}
\fontsize{6}{7.2}
\selectfont
    \caption{The training time (T.t), Sampling time (S.t), memory usage (M) and parameters (P) of the progressive strategy. FP, FS, PP, and PS denote fixed patch, fixed stride, progressive patch, and progressive stride, respectively. The combination PP \& PS is the strategy adopted in our method.} 
    \label{tab:abprogressive}
    \centering
    \begin{tabular}{c|c|c|c|c|c}
    \toprule
        Methods & PSNR $\uparrow$ & T.t (h) $\downarrow$ & S.t (s) $\downarrow$ & M (G)  $\downarrow$ & P (M) $\downarrow$  \\
        \midrule
       FP \& FS & 26.54  & 98 & 14.24 & 24 &  61\\
       FP \& PS & 26.35  & 98 & 5.43 & 24 &  61 \\
       PP \& FS & 21.83 & \underline{52} & 9.00 & \underline{6} & 61 \\
       PP \& PS & \textbf{28.04} & 55 & \underline{2.45} & 8 & 80 \\
       Ours w/o $g$ & \underline{27.04} & \textbf{52} & \textbf{2.24} & \textbf{6} & 61 \\
       \bottomrule
    \end{tabular}
\end{table}

\subsection{Ablation study}
\label{sec:abstudy}
We conduct ablation studies of P-PatchDiff on the LOL-v2-Real dataset with a resolution of $400 \times 600$. 

\textbf{Patchifying strategy. } As shown in \Cref{tab:abpatch}, we analyse the effects of four patchifying strategies, including fixed patch~\citep{weatherdiff}, multi-scale patch~\citep{MDMS}, as well as our progressive patch and progressive patch with multi-patch alignment. For a fair comparison, we use the exact same framework except for the patchifying strategy and gradually increase the number of subsets $n$ (which controls the largest patch size). The effects of $n$ on fixed-patch and multi-scale patch can be referred to \Cref{eq:weatherdiff} and \Cref{eq:mdms}, respectively. Notably, when $n=1$, two existing strategies and P-PatchDiff w/o $g$ reduce to the fixed-patch case, thus producing identical results. When using the largest $n=5$, our progressive patch achieves superior enhancement performance while being approximately $12\times$ and $35\times$ faster than the fixed-patch and multi-scale patch strategies, respectively. 

These results further indicate that simply increasing patch size within a single scale does not consistently yield optimal results, as seen in the performance drop from $n=4$ to $n=5$  in the fixed-patch setting. This underscores the importance of multi-scale information in low-light image enhancement. However, due to the lack of efficient sampling mechanisms, multi-scale patch methods incur substantial computational overhead as $n$ increases. Moreover, current multi-scale strategies fail to effectively preserve and utilise information from each scale throughout the denoising process. As a result, when $n$ is large, the influence of large patches can dominate and suppress the contributions of smaller patches, leading to suboptimal outputs. \Cref{tab:abpatch} also shows that the sampling time of both fixed-patch and multi-scale strategies grows significantly with larger $n$, whereas our P-PatchDiff maintains consistently low sampling times regardless of $n$.

Furthermore, we conducted additional experiments with $n=6$ and $n=7$ on the LOL-v2 dataset for our progressive setting. P-PatchDiff achieves 27.85/0.885/2.89 (PSNR/SSIM/sampling time) when $n=6$ and 27.92/0.884/3.17 when $n=7$, compared with 28.04/0.887/2.45 when $n=5$. These results indicate that increasing $n$ beyond 5 does not consistently improve performance in our setting. Moreover, larger $n$ leads to increased sampling time and memory consumption. Since each additional scale requires an extra diffusion step, the computational burden grows when $n$ exceeds a certain threshold. Considering the marginal performance change together with the increased computational cost, we select $n=5$ as the default configuration to achieve a better trade-off between performance and efficiency.

\textbf{Progressive strategy. } As shown in \Cref{tab:abprogressive}, we analyse the impact of the progressive strategy by training and evaluating four frameworks: fixed patch (FP) with fixed stride (FS), fixed patch with progressive stride (PS), progressive patch (PP) with fixed stride and P-PatchDiff (PP and PS). In the FP and FS setting, we use the same largest patch size ($p=192$) as P-PatchDiff and the smallest stride ($s=16$) to ensure optimal enhancement quality. Even with $p=192$, FP \& FS are still worse than P-PatchDiff due to their lack of multi-level information. The other two strategies also face challenges in training/sampling times and performance, as discussed in \Cref{sec:revisiting}. In contrast, P-PatchDiff combines PP and PS, reducing both time and memory consumption while achieving superior enhancement quality.

\begin{figure}
\centering
\includegraphics[width=1\linewidth]{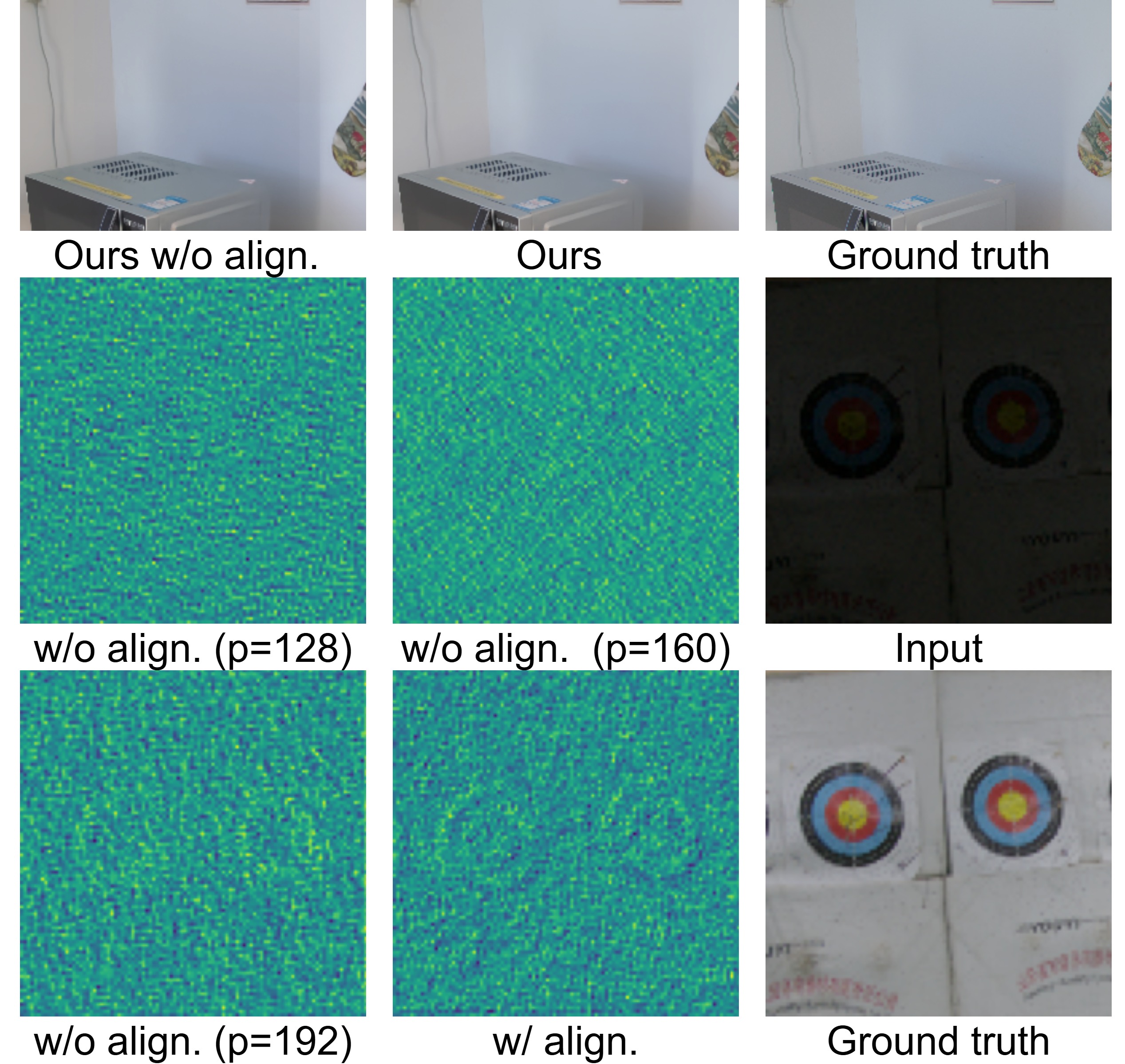}
\caption{Effects of the multi-patch alignment at the image level (top row) and feature level (bottom two rows).}
\label{fig:aligner}
\end{figure}

\textbf{Multi-patch alignment. } As shown in \Cref{tab:abstudyg}, we first evaluate the influence of multi-patch alignment on each dataset, observing a PSNR improvement of around 1 dB on three datasets. The visual results in \Cref{fig:aligner} further show that this strategy effectively  improves brightness consistency, particularly in shadow regions. In addition, the feature-level visualisations in \Cref{fig:aligner} reveal how it separates different objects in the image to facilitate denoising. Furthermore, \Cref{tab:abpatch} shows that incorporating alignment increases the sampling time by only 10\%, as it requires just \textit{a single run} per sampling step to estimate the  brightness proxy.

\begin{table} 
\fontsize{6}{7.2}
\selectfont
    \caption{Ablation study on the brightness proxy estimator ($g$) on the LOL-v1, LOL-v2-Real and the LOL-v2-Syn testing set. }
    \label{tab:abstudyg}
    \centering
    \begin{tabular}{c|c|c|c|c}
    \toprule
       \multirow{2}{*}{Methods} & \multicolumn{4}{c}{LOL-v1} \\
       \cmidrule{2-5}
         & PSNR $\uparrow$ & SSIM $\uparrow$ & LPIPS $\downarrow$ & FID $\downarrow$  \\ 
        \midrule 
        Ours w/o $g$  & 26.38 & 0.877 & 0.097 &	 50.69  \\
        Ours          & 27.18 &	0.880 &	 0.096 &	52.35  \\
        \midrule
        \multirow{2}{*}{Methods} & \multicolumn{4}{c}{LOL-v2-Real} \\
        \cmidrule{2-5}
         & PSNR $\uparrow$ & SSIM $\uparrow$ & LPIPS $\downarrow$ & FID $\downarrow$  \\ 
         \midrule
         Ours w/o $g$ & 27.04 & 0.885 &	0.142 &	 69.05  \\
         Ours         & 28.04 &	0.887 &	 0.135 &  66.02  \\
        \midrule
        \multirow{2}{*}{Methods} & \multicolumn{4}{c}{LOL-v2-Syn} \\
        \cmidrule{2-5}
         & PSNR $\uparrow$ & SSIM $\uparrow$ & LPIPS $\downarrow$ & FID $\downarrow$  \\ 
         \midrule
         Ours w/o $g$ & 26.76 & 0.933 &	0.060 &	 23.48  \\
         Ours         & 28.15 &	0.941 &	 0.054 &  23.90  \\
         \bottomrule
    \end{tabular}
\end{table}

\section{Discussion}
\subsection{Generalisation to other restoration tasks}
To further evaluate the  generalisation ability of P-PatchDiff in other scenarios, we additionally conduct experiments on two representative restoration tasks, deblurring and deraining.

\begin{table}[t]
\fontsize{6}{7.2}
\selectfont
    \caption{ Quantitative comparisons on the GoPro and Rain100L datasets for deblurring and deraining, respectively.} 
    \label{tab:abdeblurrain}
    \centering
    \begin{tabular}{c|c|c|c}
    \toprule
    Deblurring  &  Methods & PSNR $\uparrow$ & SSIM $\uparrow$  \\
        \midrule
        \multirow{3}{*}{Task-agnostic}
        & SwinIR & 24.52 & 0.773 \\
       & Restormer & \underline{27.22} & \underline{0.829} \\
      & WeatherDiff $\diamond$ & 22.16 & 0.786 \\
        \midrule
        \multirow{2}{*}{All-in-one}  
       & AirNet &  24.35 & 0.781 \\
      & Transweather &  25.12 & 0.757 \\
       \midrule
        \multirow{2}{*}{Low-light specific}  
      &  Retinexformer & 25.09 & 0.779 \\
     &  Ours $\diamond$  & \textbf{27.61} & 
       \textbf{0.886}  \\     
       \bottomrule
    \end{tabular} 
        \begin{tabular}{c|c|c|c}
    \toprule
   Deraining  &  Methods & PSNR $\uparrow$ & SSIM $\uparrow$  \\
        \midrule
        \multirow{3}{*}{Task-agnostic}
       & SwinIR &30.78 & 0.923 \\
       & Restormer & \textbf{34.81} & 0.962 \\
      & WeatherDiff $\diamond$ & 30.53 & \underline{0.970} \\
       \midrule
        \multirow{2}{*}{All-in-one}  
      &  AirNet & 32.98 & 0.951 \\ 
      & Transweather & 29.43 & 0.905 \\
       \midrule
        \multirow{2}{*}{Low-light specific}  
      &  Retinexformer & 32.68 & 0.940 \\
      & Ours $\diamond$  & \underline{34.28} & \textbf{0.972}  \\
       \bottomrule
    \end{tabular}
\end{table}

\begin{figure*}
\centering
\includegraphics[width=0.65\linewidth]{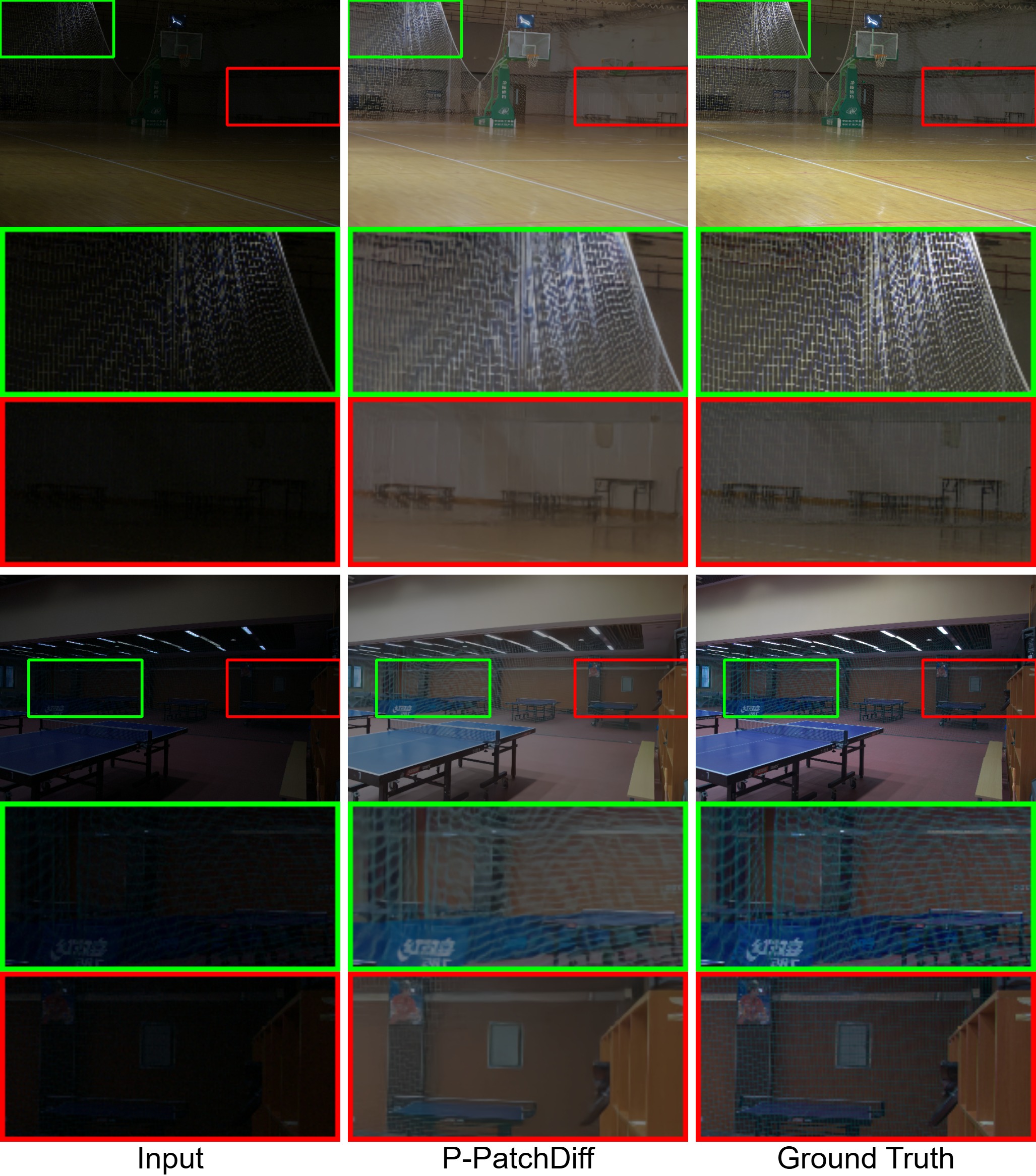}
\caption{Qualitative comparison of successful and failed cases, which are denoted by \fcolorbox{green}{white}{green} and \fcolorbox{red}{white}{red} bounding boxes, respectively. When a region is extremely dark, P-PatchDiff tends to ignore its structures (e.g., the net) entirely, rather than attempting to enhance them but failing.}
\label{fig:failure}
\end{figure*}

\textbf{Dataset. } Following prior work~\citep{Restormer}, we adopt the GoPro~\citep{nah2017deep} dataset for deblurring and the Rain100L~\citep{yang2019joint} dataset for deraining. The GoPro dataset contains $2,103$ training images and $1,111$ testing images, with a resolution of 1280 × 720. The Rain100L dataset includes 200 training images and 100 testing images, with a resolution of 480 × 320. We train and evaluate P-PatchDiff on the official training and test sets.

\textbf{Baselines. } For deblurring and deraining, we compare with task-agnostic methods (SwinIR~\citep{liang2021swinir}, Restormer~\citep{Restormer} and WeatherDiff~\citep{weatherdiff}), all-in-one restoration methods (AirNet~\citep{li2022all} and Transweather~\citep{valanarasu2022transweather}) and a low-light-specific method (Retinexformer~\citep{Retinexformer}). 

\textbf{Results. } As reported in \Cref{tab:abdeblurrain},   P-PatchDiff generalises well to both tasks, achieving competitive performance. In particular, the strong SSIM on GoPro indicates reasonable structure preservation on real-world images. On Rain100L, our method does not achieve the best PSNR, which may be attributed to the limited scale of the synthetic dataset (200 low-resolution training images), posing challenges for diffusion-based modelling.

\subsection{Failure case analysis}
As shown in \Cref{fig:failure}, we present a qualitative failure case analysis of P-PatchDiff. We observe that when a region is extremely dark, P-PatchDiff may completely omit the structures within it: as highlighted in the \fcolorbox{red}{white}{red} box, the net in front of the background entirely disappears after enhancement, with no blurry or partially recovered texture left behind.  This occurs because the input signal in such regions is too weak for the model to reliably detect the presence of structure. Interestingly, this failure is region-dependent rather than content-dependent. As shown in the \fcolorbox{green}{white}{green} box, the same net texture in a less degraded region is faithfully restored, with its fine details well preserved. This suggests that the model is inherently capable of restoring such textures. 

 We hypothesise that this behaviour also relates to the lack of semantic consistency modelling across an image. Without such modelling, the model has no higher-level understanding of the scene to compensate when local structural cues are insufficient. In contrast, the same texture in a brighter region retains enough signal for successful enhancement. We conjecture that this issue could be alleviated if the model were able to understand the semantics of the scene. With such high-level understanding of what the image depicts and how a real-world scene should be structured, the model could infer the presence of these structures from weak signals, rather than ignoring them entirely. We leave the exploration of such a semantic-consistency-aware enhancement as future work.

\subsection{Limitations}
Despite the effectiveness of P-PatchDiff, several limitations still remain and could be further explored in future work.

First, although our progressive patchifying strategy improves memory efficiency and enables flexible processing, it does not substantially reduce the overall computational cost, since we crop patches from the full-resolution image. Second, to mitigate boundary artefacts, we enforce the stride to be less than half of the patch size, which introduces duplicated pixels. While this strategy effectively reduces artefacts, it increases redundant computation in overlapping areas. Recent diffusion methods~\citep{stabledf,lightendiff} perform operations in latent space to reduce spatial resolution and achieve more efficient training and inference. Therefore, incorporating progressive patching in the latent space could further reduce computational cost and improve robustness to boundary artefacts.

\section{Conclusion}
In this paper, we presented P-PatchDiff, a scalable progressive patch diffusion model for low-light image enhancement. This progressive patch strategy offers two unique advantages over previous methods. It not only achieves competitive enhancement quality with minimum computational costs but also balances local dark-region enhancement against globally coherent brightness.  P-PatchDiff leverages a novel progressive training and sampling strategy that gradually shifts from local to global views, capturing multi-level information for effective local dark-region enhancement without sacrificing training or sampling efficiency. Additionally, we introduced a multi-patch alignment strategy that leverages an estimated global brightness proxy, resolving the output inconsistencies that arise when diffusion models use varying patch sizes. By unifying local enhancement and global brightness alignment, our method effectively handles spatially variant low-light images. Extensive experiments on ten low-light image enhancement datasets demonstrate the effectiveness and efficiency of our approach. In the future, we will continue to explore efficient diffusion models for low-light image enhancement.






\section*{Declarations}

\begin{itemize}
\item Data availability: All datasets used in this study are publicly available. The corresponding dataset papers have been cited in the manuscript.
\item Code availability: The code is available at \url{https://github.com/RuoyuGuo/P-PatchDiff}.
\end{itemize}

\bibliography{sn-bibliography}

\end{document}